\documentclass[pdflatex,sn-mathphys-num,Numbered]{sn-jnl}

\usepackage{graphicx}%
\usepackage{multirow}%
\usepackage{amsmath,amssymb,amsfonts}%
\usepackage{amsthm}%
\usepackage{mathrsfs}%
\usepackage[title]{appendix}%
\usepackage{xcolor}%
\usepackage{textcomp}%
\usepackage{manyfoot}%
\usepackage{booktabs}%
\usepackage{algorithm}%
\usepackage{algorithmicx}%
\usepackage{algpseudocode}%
\usepackage{listings}%
\usepackage[caption=false]{subfig}

\usepackage{siunitx}
\makeatletter
\gdef\orcidlogo{} 
\makeatother

\begin{document}
\title{RESONATE: a Real Social Navigation Testbed \\ for Autonomous Service Robots}

\author[1]{\fnm{Stefano} \sur{Trepella}}\email{stefano.trepella@polito.it} 
\equalcont{These authors contributed equally to this work.}

\author*[1]{\fnm{Andrea} \sur{Ostuni}}\email{andrea.ostuni@polito.it} 
\equalcont{These authors contributed equally to this work.}

\author[1]{\fnm{Mauro} \sur{Martini}}\email{mauro.martini@polito.it} 
\author[2]{\fnm{Irene} \sur{Borgini}}\email{irene.borgini@intesasanpaolo.com} 
\author[1]{\fnm{Marcello} \sur{Chiaberge}}\email{marcello.chiaberge@polito.it} 

\affil*[1]{\orgdiv{Department of Electronics and Telecommunications}, \orgname{Politecnico di Torino}, \orgaddress{\street{Corso Duca degli Abruzzi 24}, \city{Torino}, \postcode{10129}, \state{Torino}, \country{Italy}}}

\affil[2]{\orgdiv{Innovation Center}, \orgname{Intesa Sanpaolo}, \orgaddress{\street{Corso Inghilterra 3}, \city{Torino}, \postcode{10138}, \state{Torino}, \country{Italy}}}


\abstract{Social navigation is essential for autonomous service robots operating in human environments, but progress remains constrained by the lack of extensive real-world benchmarks. This paper presents the Real Social Navigation Testbed (RESONATE), a reproducible benchmark for human-aware navigation based on a unified robotic platform, a shared ROS~2 software stack, precise trajectory ground truth, and structured social interaction scenarios. We evaluate eight controller configurations, combining classical and predictive local planners with different social modeling strategies, over eight representative scenarios with five repetitions each, yielding 320 physical experiments. The benchmark combines quantitative navigation and social metrics with participant-based qualitative assessments to jointly measure efficiency, safety, comfort, and perceived social behavior. Results reveal scenario-dependent trade-offs among the tested methods. The algorithms that stands out for their performance are the DWA with Gaussian social costmaps for its reliability, and the SFM-MPPI to provide the best balance between robustness, navigation efficiency, and social acceptance. Overall, RESONATE supports the robotics community with a real-world benchmark, reusable baselines, data, and testbed, and a consistent evaluation protocol for advancing socially aware navigation research.}




\maketitle

\section{Introduction}
In recent years, human-aware navigation, also known as social navigation, has emerged as a major research direction in mobile service robotics. In these settings, the robot must not only reach its goal safely and efficiently, but also move in ways that are comfortable, legible, and socially acceptable to nearby people. The study and development of social navigation techniques are crucial, as autonomous mobile robots are moving from controlled industrial settings to crowded indoor environments. Relevant applications include sanitization, hospital logistics, and airport mobility assistance~\cite{huang2023hospitals, Blesson2023airportwheelchair}, all of which depend on navigation behaviors that coexist with human traffic while respecting distances and social conventions.

The possible benefits of human-aware navigation make it fundamental to develop efficient and reliable control strategies. Accordingly, the community has proposed a broad range of solutions, from socially compliant motion models learned from demonstrations to planners that explicitly aim to reach interpretable robot motion \cite{Mavrogiannis2023}. However, the advancement of research remains unclear due to the difficulty of fairly comparing the studies. Reported performance often depends not only on the navigation algorithm itself, but also on the adopted sensing pipeline, controller implementation, social model, computational budget, and scenario design.

Evaluation is therefore one of the main unresolved challenges in social navigation. Recent reviews have highlighted the lack of common protocols, metrics, and benchmarks, especially when navigation criteria (trajectory efficiency) and human-centered criteria (safety, naturalness, social discomfort) must be considered jointly \cite{gao2022evaluation,Mavrogiannis2023,Francis2025guidelines}.
Important recent efforts have addressed parts of this problem through virtual benchmarks, simulators, datasets, and sparse real-world evaluations.
SocNavBench~\cite{biswas2022socnavbench} and HuNavSim~\cite{PerezRal2023} are valuable examples of simulation-based benchmarking tools that support repeatable testing and fast comparison of alternative navigation methods. SocNavBench, however, is based on a pre-collected set of trajectories and does not fully support human reaction to the robot.
On the other hand, SCAND~\cite{karnan2022scand} provides a dataset of real trajectories to boost imitation learning approaches. This can be considered a valuable resource to train policies or cost models. Nonetheless, it cannot serve as a reference benchmark for testing relevant algorithms.
Kobayashi et al. \cite{kobayashi2022robot} validated their navigation method in a controlled environment using a differential-drive robot to evaluate collision rates and social interference. Hirose et al. \cite{hirose2024selfi} evaluated their SELFI framework across three distinct real-world environments, utilizing reproducible settings and a qualitative user study to measure adherence to social norms. Such experimental validation attempts remain limited to singular studies, whereas it would be beneficial for the research community to have access to shared baselines implementations, metrics, and protocols. These resources will uniform real validation testing and boost research novelty production.
Two indoor human-robot interaction datasets, TH\"OR~\cite{rudenko2020thor} and TH\"OR-MAGNI~\cite{schreiter2025thormagni}, rely on multi-camera motion capture to record accurate trajectories of people and mobile robots across structured scenarios. However, they are designed for the observational analysis and prediction of human motion around a robot, rather than for the closed-loop comparison of navigation controllers. Conversely, ROS-based benchmarking suites such as SEAN~2.0~\cite{tsoi2022sean2} and Arena-Bench~\cite{kastner2022arenabench} formalize canonical social situations and evaluate local planners under common metrics, but they do so in simulation, where pedestrians react to the robot according to behavior models rather than as real people.
Nevertheless, the literature still lacks an extensive real-world benchmark that compares multiple social navigation configurations under a unified hardware platform, a shared software stack, precise trajectory ground truth, repeated trials, and a structured set of canonical interaction scenarios.

To address this gap, we present RESONATE, a real-world benchmark for social navigation with autonomous service robots. RESONATE evaluates widely used and recent local planning solutions through a precise ground-truth acquisition system and a set of categorized interaction scenarios. The benchmark comprises eight controller configurations built from two base local planners (DWA and MPPI) implementing across three social cost variations (obstacle-only baseline, Gaussian social costmap, and SFM-based social work cost), alongside two standalone crowd-model controllers (ORCA and SFM).
These configurations are evaluated across eight scenarios, including passing, crossing, and more complex encounters, with five repetitions per condition, for a total of 320 physical experiments.Unlike the datasets and simulation suites discussed above, RESONATE combines motion-capture ground truth of people and robot in structured scenarios with a comparison of local planners in canonical social situations through a shared ROS~2 and Nav2 stack, run in closed loop with a physical robot and real participants who also rate each run. To promote reproducibility and practical relevance, all experiments are conducted on a widely adopted service-robot platform, the Clearpath Jackal, and implemented within a ROS~2 and Nav2-based software pipeline.
Beyond the extensive benchmark itself, this work provides a reusable experimental asset for the community. We publicly release the modular ROS~2 and Nav2 software plugins for integrating different social costs into navigation controllers, together with the dataset of collected trajectories with precise ground truth and the associated participant evaluations (see the Data and Code availability statements).
Overall, RESONATE provides a real-world testbed and a consistent experimental protocol to advance future research in social navigation.

The paper is organized in the following sections: Section~\ref{sec:methodology} outlines the adopted evaluation setup, detailing the benchmark structure (\ref{subsec:res_benchmark}), along with the specific hardware (\ref{subsec:hardware_setup}) software configurations (\ref{sec:software_setup}), participant surveys (\ref{subsec:survey}) and tested algorithms (\ref{subsec:algorithms}) used throughout the experimental process.
Section~\ref{sec:benchmark_metrics} focuses instead on the metrics used for evaluating the social navigation strategies: in \ref{subsec:quantitative_metrics} the metrics obtained by elaborating the quantitative results are explained in detail, while \ref{subsec:qualitative_metrics} is focused on the explanation of the metrics obtained by the surveys conducted on the participants. 
 Section~\ref{sec:exp_evaluation} describes the experimental evaluation and scenario configurations. Section~\ref{sec:results} presents and discusses the quantitative, proxemics, and human evaluation results.
\section{Methodology}
\label{sec:methodology}
This section describes the hardware and software platform. The complete setup used for the benchmarking features includes commercially available solutions for both the robot and the sensors employed. The libraries used have been integrated into ROS~2 to facilitate the replicability of the described scenarios in different environments.
\subsection{Hardware Platform} \label{subsec:hardware_setup}
\begin{figure}[htbp]
    \centering
    \subfloat[]{%
        \includegraphics[width=0.48\linewidth]{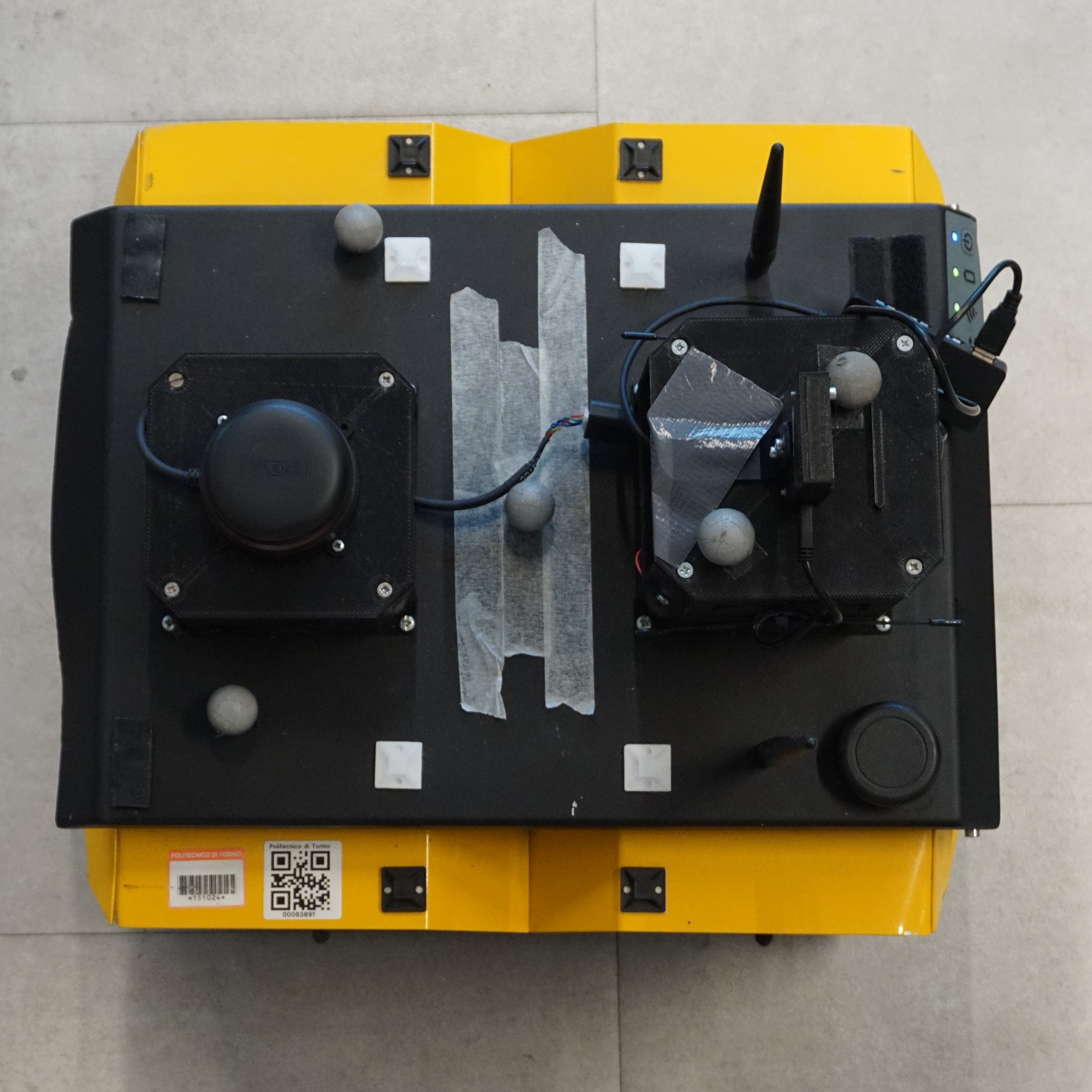}%
        \label{fig:jack_top}%
    }
    \hfill
    \subfloat[]{%
        \includegraphics[width=0.48\linewidth]{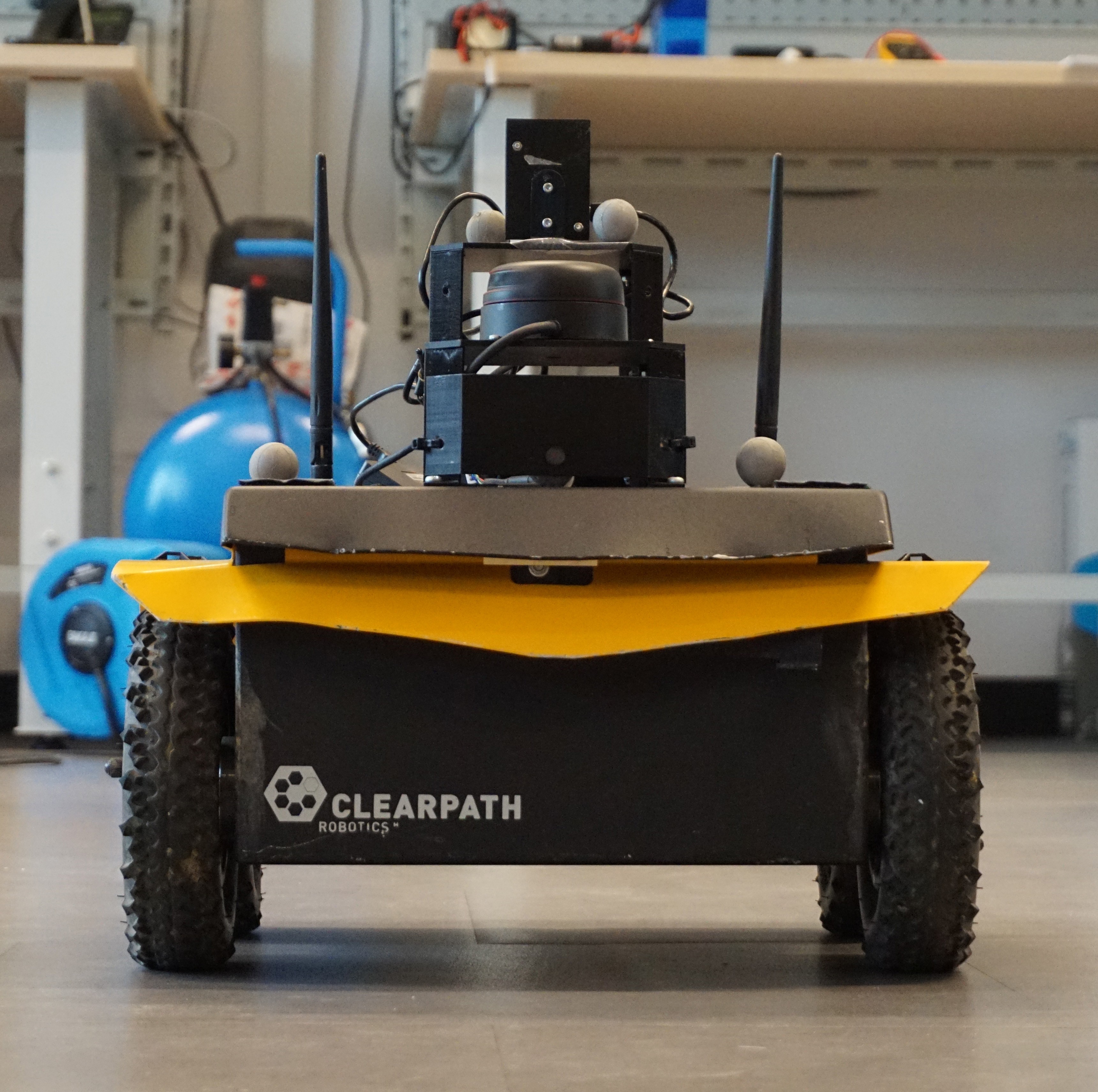}%
        \label{fig:jack_front}%
    }

    \vspace{0.2em}

    \subfloat[]{%
        \includegraphics[width=0.48\linewidth]{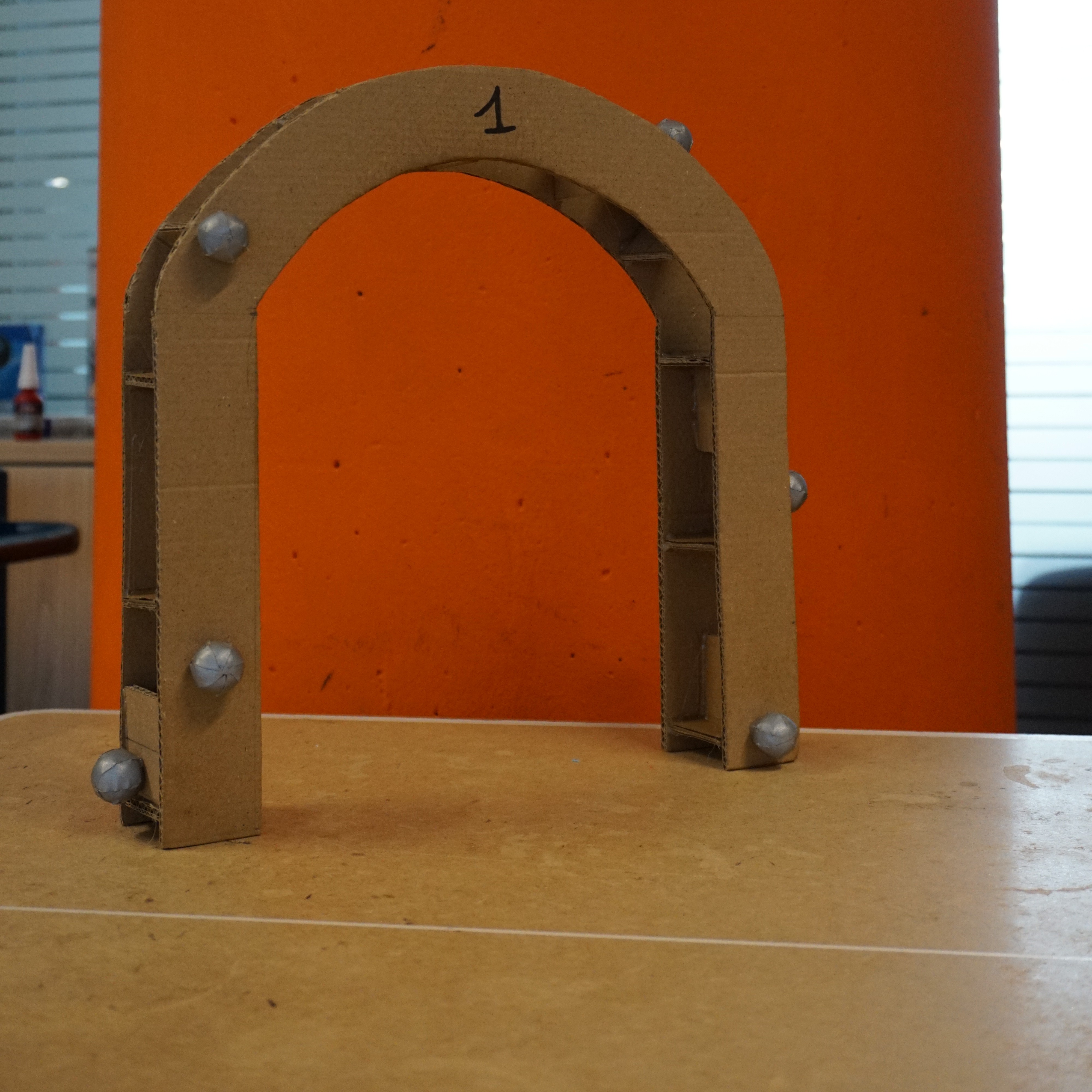}%
        \label{fig:ag_collar}%
    }
    \hfill
    \subfloat[]{%
        \includegraphics[width=0.48\linewidth]{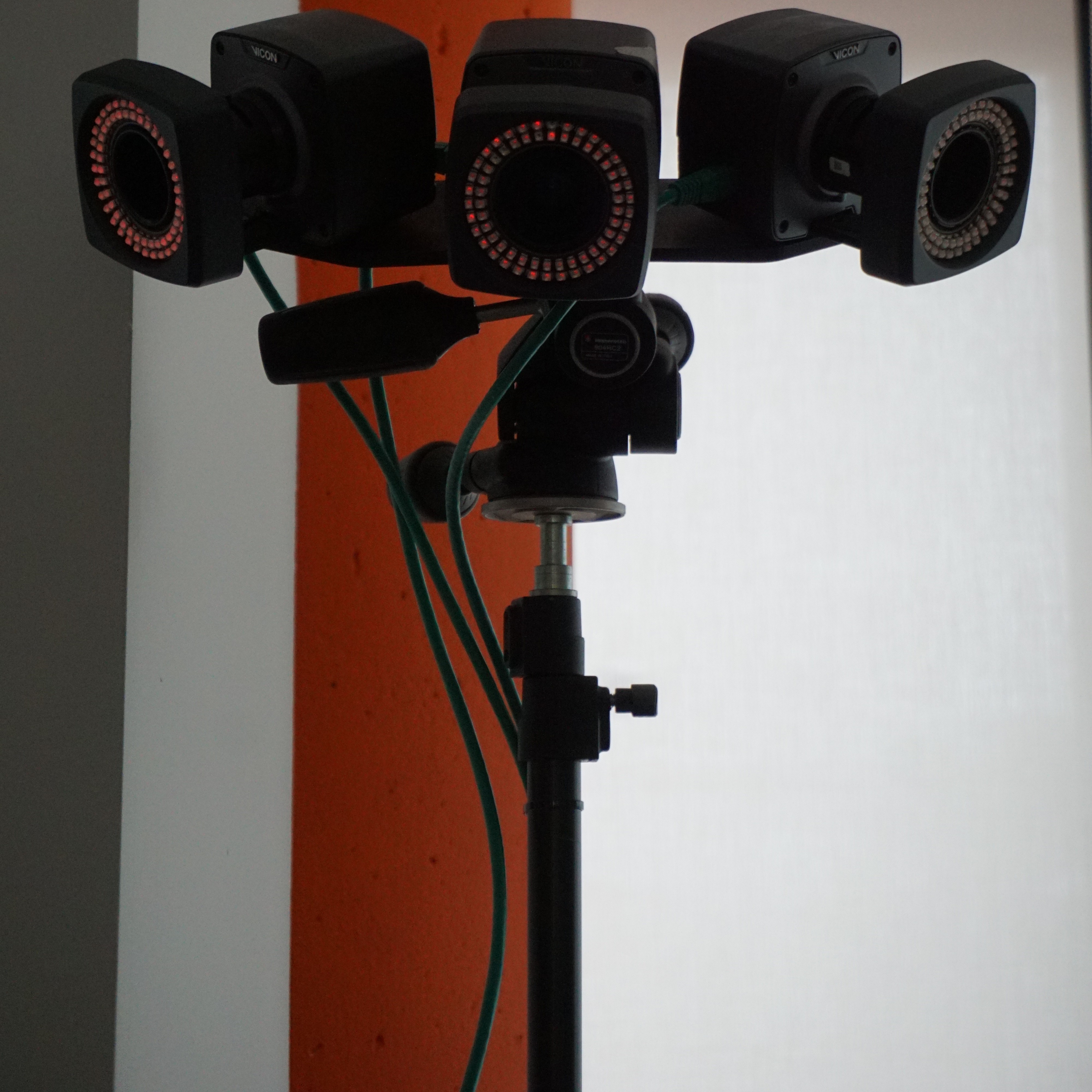}%
        \label{fig:vic_cam}%
    }

    \caption{The hardware used including: Jackal (\ref{fig:jack_top} and \ref{fig:jack_front}), agent neck-worn tracking frame (\ref{fig:ag_collar}) and VICON cameras (\ref{fig:vic_cam}).}
    \label{fig:hardware}
\end{figure}

The robot used across the different scenarios is the Jackal from ClearPath, a commercial platform used in research, equipped with an Intel i7 6700 processor and an RPLidar 2D LiDAR, used for obstacle detection. People pose tracking was handled using a motion capture system built with ten VICON cameras. The cameras were allowed to record the positions of every agent, including the robot, with a millimetric error. Every participant was given a U-shaped \qtyproduct{33 x 30}{\centi\meter} neck-worn device to make them recognizable by the tracking system. Each wearable device featured a unique arrangement of spheres surrounded by reflective tape to uniquely identify each person. The robot localization was also handled using the VICON cameras. The hardware is shown in Fig. \ref{fig:hardware}. Fig. \ref{fig:jack_top} and \ref{fig:jack_front} show the rover, Fig. \ref{fig:ag_collar} shows an agent tracking device, and Fig. \ref{fig:vic_cam} shows some of the cameras used.
The effective experimental area is \qtyproduct{4.0 x 5.0}{\meter}, although not all experiments employed the same map.
The velocity limits of the robot have been set to $v \in [0.0, 0.6]$ \unit{\meter\per\second} and $\omega\in [-1.5, 1.5]$ \unit{\radian\per\second} for all experiments and the different controllers used.

\subsection{Benchmarking Software Architecture} \label{sec:software_setup}
The controllers were implemented and tested using the ROS~2 Humble framework \cite{ros2}, featuring the Nav2 navigation stack \cite{macenski2023survey}. Integrating the local controllers into Nav2 enables a fair assessment of performance differences across approaches. The other elements and associated parameters of the navigation stack, such as the global planner (NavFn) and the behavior tree, were kept constant. The plugins tested were integrated as C++ libraries. Standard Python launch files have been used to start the entire navigation and data collection system: relay VICON messages containing the poses of the involved elements to the Jackal, set the Nav2 goal, and start recording the ROS~2 bag for each experiment. The information from each bag was then ready to be extracted, filtered, and processed using an offline evaluation pipeline.



\begin{figure*}[t]
    \centering
    \includegraphics[width=\linewidth]{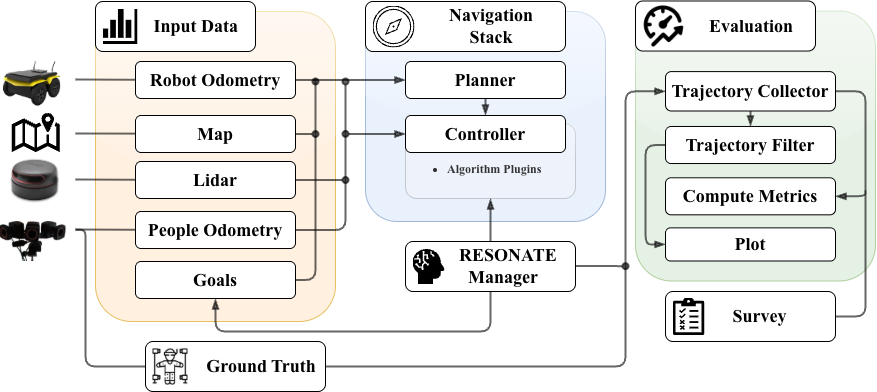}
    \caption{The schematics for the RESONATE benchmark. The three blocks of the RESONATE benchmark suite are the input data, the navigation stack, and the evaluation, divided into the online trajectory collection and offline filtering and evaluation. The survey is acquired separately.}
    \label{fig:resonate_benchmark_schematics}
\end{figure*}

\subsection{RESONATE Benchmarking Suite} \label{subsec:res_benchmark}
A summary of the hardware and software components used for conducting the benchmark is presented in Fig. \ref{fig:resonate_benchmark_schematics}.
The RESONATE manager is the software pivotal block of the system: it sets up the entire navigation stack providing input data, parameters, costs, and controller plugins, and it coordinates the starting time of the trajectory collection.
The first block of the entire pipeline (yellow block on the left in Fig. \ref{fig:resonate_benchmark_schematics}) is composed of the input data, necessary for the navigation stack to execute correctly. It is composed of both robot and people odometry obtained via the multi-camera tracking system, along with the goal pose requested, the obstacle points sensed by the LiDAR mounted on the robot at a height of 35 \unit{\centi\meter} from ground, and the environment map. These data feed into the second block, represented by the navigation stack (blue central block in Fig. \ref{fig:resonate_benchmark_schematics}), that is composed mainly of the global and local planners, and by the map and cost nodes. The local controller is represented by one of the algorithms used in the benchmark comparison. At the start of each run, the RESONATE manager sets the goal defined for each scenario for the navigation stack and starts the trajectory recording, which is part of the evaluation block (green block on the right in Fig. \ref{fig:resonate_benchmark_schematics}). The positions and velocities of both the robot and the agents are saved during the run using VICON cameras, enabling the minimization of localization and perception errors in the ground truth. The positions are then filtered offline to remove residual noise, and they are used to compute the quantitative metric values.
The data for qualitative metrics are instead collected from participants via a survey, as detailed in Sec. \ref{subsec:survey}.
The experiments have one of the following possible outcomes:
\begin{itemize}
    \item Success: the robot reached its goal without collisions.
    \item Collision: the robot collided with the environment or another agent.
    \item Timeout: the robot was not able to reach its goal in a reasonable time (we set the timeout at 60 seconds).
    \item Navigation Failure: the robot failed to reach its goal due to a faulty behavior of the navigation stack.
\end{itemize}

Path length, time to goal, and heading changes are computed only on successful runs, since they depend on the robot reaching its goal and can otherwise be altered by early failures. The social metrics ($\mathrm{SW_{s}}$, AMD and PR) are computed on all runs, since the robot interacts with the agents regardless of the outcome.

\subsection{Survey}\label{subsec:survey}
30 different people participated in the experiments and were asked to evaluate them at the end of the experience. The population of participants has been selected from a pool of Italian mixed gender volunteers (76\unit{\percent} male and 24\unit{\percent} female), all aged 20-40 years old, with an average age of 27 and a standard deviation of 5 years. The majority of the participants reported that they had some familiarity with robots using a 5-point Likert scale. The number of available participants allowed us to create unique combinations of people for each run and scenario. At the end of the experiment, each participant was given a survey to evaluate the tested algorithms. The specific controller used in the experiment was not indicated to avoid bias in the evaluation. The metrics used in the survey are described in Section \ref{subsec:qualitative_metrics} and were evaluated on a five-point Likert scale. Participants were asked to grade the qualitative metrics regardless of the result, since the robot always interacts with the participants. 

\subsection{Reproducibility}\label{sec:reproducibility}
Reproducibility is a key aspect when conducting a benchmark; we tried to ensure it by considering these aspects:
\begin{itemize}
    \item \textit{Commercially available hardware}.\\ The only custom pieces are the agents' neck-worn tracking frames, which can be easily reproduced using cardboard and reflective tape. It is fundamental that retro-reflective markers are placed in a unique pattern for each agent, to ensure that each person can be tracked smoothly.
    \item \textit{Open-source software framework based on Nav2}.\\ All the tested algorithms are modular and usable by leveraging standard and reliable ROS~2 packages, eliminating the need for custom setups and allowing for the possibility of expanding the benchmark in the future by adding other controllers. The developed critics are seamlessly integrated into Nav2 as standard plugins. Additional configuration files are provided to reproduce the exact Nav2 settings.
    \item \textit{Easily replicable scenarios}. \\ The experiments are small-scale, and little space is required to set up the described scenarios.
\end{itemize}
The maps, plugins, nodes, and configuration files used for navigation are available in the RESONATE repository\footnote{\url{https://github.com/PIC4SeR/RESONATE}}; installation instructions are given in Online Resource 1.
We adopt ROS~2 middleware, not ROS, since the latter reached End of Life in January 2025. We decided to test only algorithms available for ROS~2 to improve reproducibility, using a software platform that is still being updated.

\subsection{Algorithms}\label{subsec:algorithms}
In our work, given the variety of different approaches for social robot navigation we selected a representative set of planners that span the frequently used methodological frameworks. In particular, we include a set of algorithms to capture different assumptions on human-robot interaction, planning structure, and control strategies. The selected methods have also been chosen based on the availability of open-source implementations, their adoption in the community, and their compatibility with real-time robotic systems. We decided to employ approaches that could be run in real time on the onboard Jackal computer, which could handle the chosen algorithms at a consistent frequency of \qty{20}{\Hz}. 


\begin{itemize}
    \item The \textbf{Dynamic Window Approach} (DWA) \cite{fox1997dynamic} serves as a classic local planner widely adopted for obstacle avoidance in human-aware navigation, relying on heuristic cost functions over velocity samples.
    \item The \textbf{Model Predictive Path Integral} (MPPI) \cite{williams2017model} is a more recent model predictive control algorithm that extends this paradigm by formulating navigation as a stochastic optimal control problem, allowing more expressive cost representations and better handling of system dynamics. 
    \item The \textbf{Optimal Reciprocal Collision Avoidance} (ORCA) \cite{ORCA_paper} represents a purely reactive geometric approach, providing a strong baseline for collision avoidance without explicit modeling of social behavior. It is widely used in simulators to model crowd motion.
    \item The \textbf{Social Force Model} (SFM) \cite{Helbing1995SocialFM} models crowd interactions as a set of forces. Agents are affected by attractive forces toward the goal and repulsive forces from the obstacles and other agents. These forces can then be used to compute the agents' future velocities. 
\end{itemize}

\begin{figure}
    \centering
    \subfloat[Gaussian Social Costmap\label{fig:gaussian_costmap}]{
        \includegraphics[width=\linewidth,height=7cm,keepaspectratio]{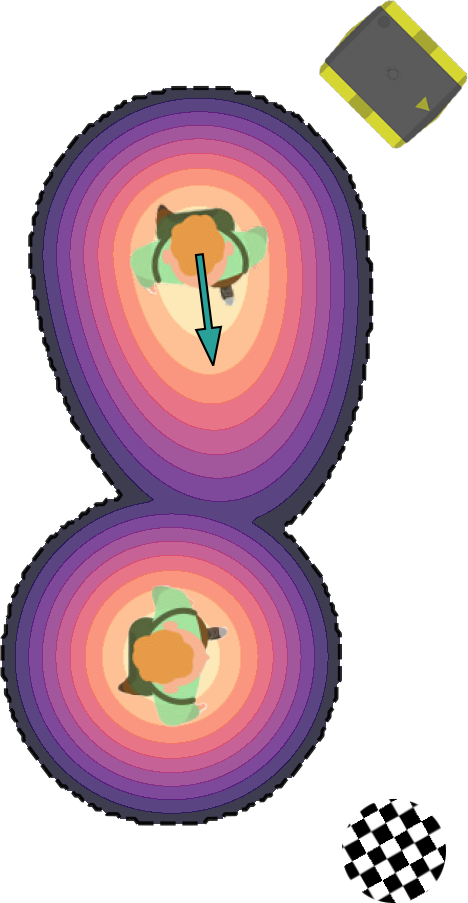}
    }\hfill
    \subfloat[Social Force Model\label{fig:social_cost}]{
        \includegraphics[width=\linewidth,height=7cm,keepaspectratio]{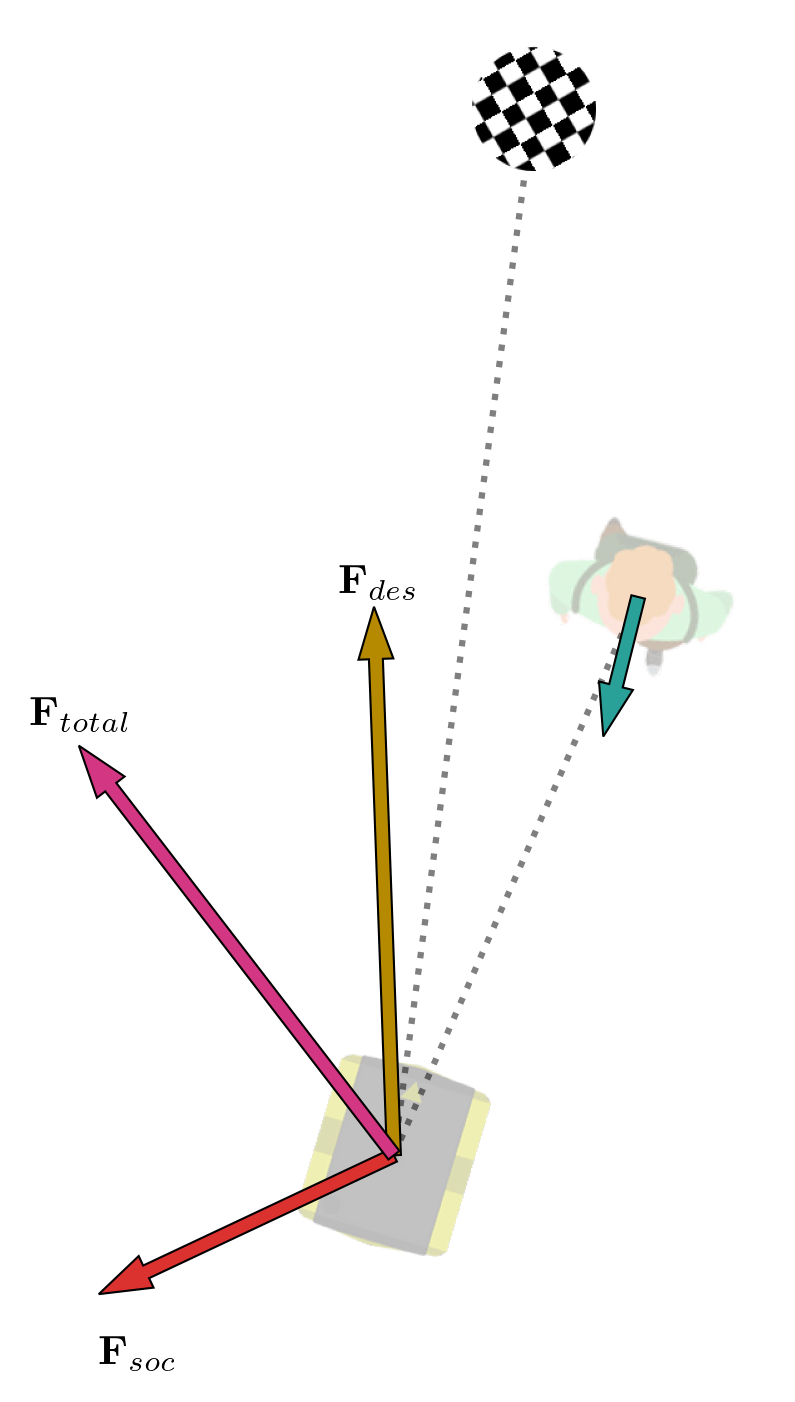}
    }
    \caption{The figure on the left, Fig. \ref{fig:gaussian_costmap}, represents how the costmap is modified by the Gaussian Costmap layer plugin. On the right, Fig. \ref{fig:social_cost} depicts the forces derived by the SFM, which will be employed to compute the Social Work.}
    \label{fig:employed_costs}
\end{figure}

All of these algorithms were used as standalone local planners, but for DWA and MPPI, three distinct variations were tested using two distinct social cost modules: a Gaussian Costmap and a Social Work critic based on the SFM.
The Gaussian Costmap directly modifies the costmap's shape around the agent, increasing cost in the region in front of the agent and discouraging the robot from crossing its path.
The Social Work critic, instead, adds a cost that is the cumulative sum of the norms of the social forces at each time step. Minimizing this additional cost should reduce hazardous trajectories and lead to more socially aware behaviors. 
These modules were not implemented in the ORCA and SFM, since they already embed a model of the agents' behavior. These plugins are represented in Fig. \ref{fig:employed_costs}.

Other promising approaches present in the literature, such as \cite{SICNav-Diffusion}, were not provided in a publicly available ROS~2 version or were not feasible to run on the same hardware as the other algorithms. 
The benchmark was originally intended to include state-of-the-art methods in reinforcement learning (RL). However, most of the RL policies trained with social costs, as done in \cite{CADRL_paper}, proved to be weak against real-world generalization, affecting the success of the navigation in most of the cases. Among the publicly available models, the work proposed by Chen et al. \cite{sarl_crowd_paper}, which stands out for its popularity in the community, runs at only \qty{0.5}{Hz} on the robot's computational hardware, preventing sufficiently safe and smooth navigation to be included in the benchmark.


\section{Benchmark Metrics}\label{sec:benchmark_metrics}
The choice of evaluation metrics is still an open research problem in social navigation, given the complexity of combining human comfort and trajectory optimization.
The metrics chosen for this benchmark encompass both quantitative and qualitative measures, selected in line with the most recent guidelines, works, and evaluation tools in the literature \cite{Francis2025guidelines, PerezRal2023}, providing a comprehensive view of social navigation performance. Using both types of metrics can provide deeper insights into the problem and foster research on evaluation protocols and the interpretation of metrics.

\subsection{Quantitative Metrics}\label{subsec:quantitative_metrics}
The quantitative metrics are a set of ten numerical metrics used in the literature and compiled in standard social navigation benchmarks, such as HuNavSim \cite{PerezRal2023}. We adopt four metrics focused on classic navigation efficiency, and six metrics for social compliance. The employed quantitative metrics are those presented in Table \ref{tab:quantitative_metrics_explanation}. The success rate is computed across every experimental run; PL, TTG, and CHC are averaged across successful runs, while $\mathrm{SW_{s}}$, AMD and PR are averaged across all runs.
\begin{table}
    \caption{Selected quantitative metrics}
    \label{tab:quantitative_metrics_explanation}
    \begin{tabular}{p{3.4cm}p{1.5cm}p{4.4cm}}
    \toprule
    \textbf{Complete Name} & \textbf{Acronym} & \textbf{Brief Description} \\
    \midrule
    \multicolumn{3}{l}{\textbf{Navigation metrics}} \\
    \midrule
    Success Rate & SR & Computed as the percentage of successful runs. \\
    Path Length [\unit{\meter}] & PL & Total length of the robot's path in meters. \\
    Heading Changes [\unit{\radian}] & CHC & Cumulative heading change during the complete trajectory in radians. \\
    Time to Goal [\unit{\second}] & TTG & Time taken in seconds to reach the goal. \\
    \midrule
    \multicolumn{3}{l}{\textbf{Social metrics}} \\
    \midrule
    Average Minimum Distance to Closest Agent [\unit{\meter}] & AMD & Average of the minimum distances to the closest agent throughout the run. \\
    Social Work per Second & $\mathrm{SW_{s}}$ & Social work exerted by the robot toward agents, modeled using the Social Force Model and normalized by the duration of the run. \\
    Proxemics Metrics & $\mathrm{PR_{I}}$, $\mathrm{PR_{PE}}$, $\mathrm{PR_{PU}}$, $\mathrm{PR_{S}}$ & Percentage of time the robot spent in predefined zones (public, personal, social, intimate) around agents. \\
    \bottomrule
    \end{tabular}
\end{table}
The success rate (SR), path length (PL), and time-to-goal (TTG) are strictly related to navigation performance and, sometimes, they can also support or correlate with social and overall evaluation of the experiment.
Safety and discomfort are the primary aspects to be assessed by a social navigator; hence, we included a measure of the average minimum distance (AMD) from the robot to the closest person that the robot interacted with in the experiments in the metrics set. The distance has been evaluated as  boundary-to-boundary (accounting for the Jackal's footprint and human radius).
A more advanced analysis of reciprocal distances and intrusion levels is proposed through a proxemics study. The concept of proxemics has been introduced by Hall \cite{hall1966hidden}, and it is still widely employed in social navigation studies. In this work, the proxemics metrics (PR) indicate the percentage of time that the robot spent in a specific area around the closest person throughout the experiment. The theory defines four different zones: an intimate zone (less than 0.45 meters), a personal zone (0.45 to 1.2 meters), a social zone distance typical of human interaction (1.2 to 3.6 meters), and lastly, the public zone where an agent's motion is expected to minimize discomfort on other subjects (more than 3.6 meters). The Social Work $\mathrm{SW}$ is a different quantitative metric based on the Social Force Model. It aims to quantify the total social impact the robot produces on the agents and vice versa. According to the model, it is obtained as the sum of the norms of the social forces throughout the entire trajectory of the robot; it is then normalized by the duration of the experiment to obtain the Social Work per second $\mathrm{SW_{s}}$.
Finally, we included the cumulative heading changes (CHC) as a quantitative score of the smoothness of the robot trajectory, which could affect the robot predictability and the human comfort.

\subsection{Qualitative Metrics}
\label{subsec:qualitative_metrics}
The qualitative metrics comprise a set of four human-centric metrics used to rank the experiments across relevant aspects, as proposed by the recent work of Hirose et al. \cite{hirose2024selfi}, and are compliant with the guidelines outlined in \cite{Mavrogiannis2023, Francis2025guidelines}.
The qualitative metrics are presented in Table \ref{tab:qualitative metrics}.

\begin{table}[t]
    \centering
     \caption{Selected qualitative metrics}
     \label{tab:qualitative metrics}
    \begin{tabular}{llp{4cm}}
    \toprule
    \textbf{Complete Name} & \textbf{Acronym} & \textbf{Brief description} \\
    \midrule
    Safety & SA & The ability to avoid collision, near-collision, or intruding human personal space \\
    Friendliness & FR & The compliance with expected social norms \\
    Naturalness & NA & Degree of smoothness of the robot trajectory \\
    Avoidance foresight & AF & The ability to predict a human's trajectory in advance and  respond to unexpected changes to keep safety distances \\
    \bottomrule
         
    \end{tabular}
\end{table}

Safety (SA) and friendliness (FR) represent the primary social compliance requirements for the robot, ensuring, respectively, safe, non-intrusive navigation and compliance with expected social conventions, such as keeping to the right when traversing corridors and avoiding cutting in front of people crossing. 
Naturalness (NA) and avoidance foresight (AF) are good indicators of movement quality, assessing, respectively, the smoothness of the trajectory and the planner's ability to understand and anticipate human trajectories in advance or to react promptly to abrupt changes to avoid collisions.

\section{Experimental Evaluation}
\label{sec:exp_evaluation}
\subsection{Algorithm Selection and Configuration}
\label{subsec:algo_selection}

Based on the algorithm overview in Section \ref{sec:methodology}, we tested eight different configurations of local planners. The configurations are described in Table \ref{tab:algorithms_config}.
\begin{table}[ht]
    \centering
    \caption{Algorithm description}
    \label{tab:algorithms_config}
    \begin{tabular}{ll}
    \toprule
     \textbf{Name}  & \textbf{Description}  \\
     \midrule
        DWA & Only local obstacles detection \\

        $\mathrm{DWA_{GSC}}$ & Gaussian costmap around people
        \\
    
        SFM-DWA & SFM-based social cost 
        \\
     
        MPPI & Only local obstacles detection
        \\
       
        $\mathrm{MPPI_{GSC}}$ & Gaussian costmap around people
        \\
       
        SFM-MPPI & SFM-based social cost
        \\
        
        ORCA & ORCA computes the command directly
        \\
        
        SFM & SFM computes the command directly
        \\
        \bottomrule
    \end{tabular}
\end{table}
\\
This set of algorithms and configurations allows us to make different strategic performance comparisons, never carefully underlined before:
\begin{itemize}
    \item Different families of popular local planning strategies are compared in their natural implementation (DWA and MPPI);
    \item Crowd models such as SFM and ORCA are used directly for robot commands computations and compared with other controllers;
    \item The same algorithms (DWA or MPPI) have been tested in three different configurations adopting social cost plugins: basic version, with Gaussian social costmap, and with social work cost.
\end{itemize}

A parameter tuning phase has been conducted for each controller to ensure that all algorithm implementations could operate under optimal conditions.

\begin{table*}[b]
\centering
\caption{Selected scenarios}
\label{tab:scenarios_description}
\begin{tabular}{lcp{0.5\textwidth}}
\toprule
\textbf{Name} & \textbf{Difficulty} & \textbf{Description}\\
\midrule

S1 - Passing & Moderate & The robot traverses a hallway starting from the center, and one person moves in the opposite direction, keeping to the right-hand side. \\

S2 - Overtaking & Moderate & The robot traverses a hallway starting from the right-hand side, with a person starting ahead of it and going in the same direction. \\

S3 - Crossing 1 & Moderate & The robot has to reach a goal in front of it while two people cross from the same side, with a slight delay (the second person starts crossing when the other has almost reached the goal). \\

S4 - Crossing 2 & Moderate & This crossing variation is similar, but the second person is on the opposite side at the start. \\

\midrule 

S5 - Crossing 3 & Hard & The robot traverses the room while three people cross in an X shape, two of them side-by-side. \\

S6 - Narrow turn & Hard & The robot has to turn to enter a passage while a person has to exit that same passage. The robot then has to reach the goal near another person that is walking around a fixed point. \\

S7 - Mixed & Hard & The robot has to traverse a room populated by three people, one stationary, one crosses the robot's trajectory near the start, the other passes the robot near the goal. \\

S8 - Curious person & Hard & The robot traverses the room while an agent actively tries to block its path. The person is instructed to track the robot's position to block its path while keeping a distance of about 50 centimeters. \\

\bottomrule \\

\end{tabular}
\end{table*}

\begin{figure}[t]
    \centering
    \subfloat[S1 - Passing]{
        \includegraphics[width=0.45\linewidth]{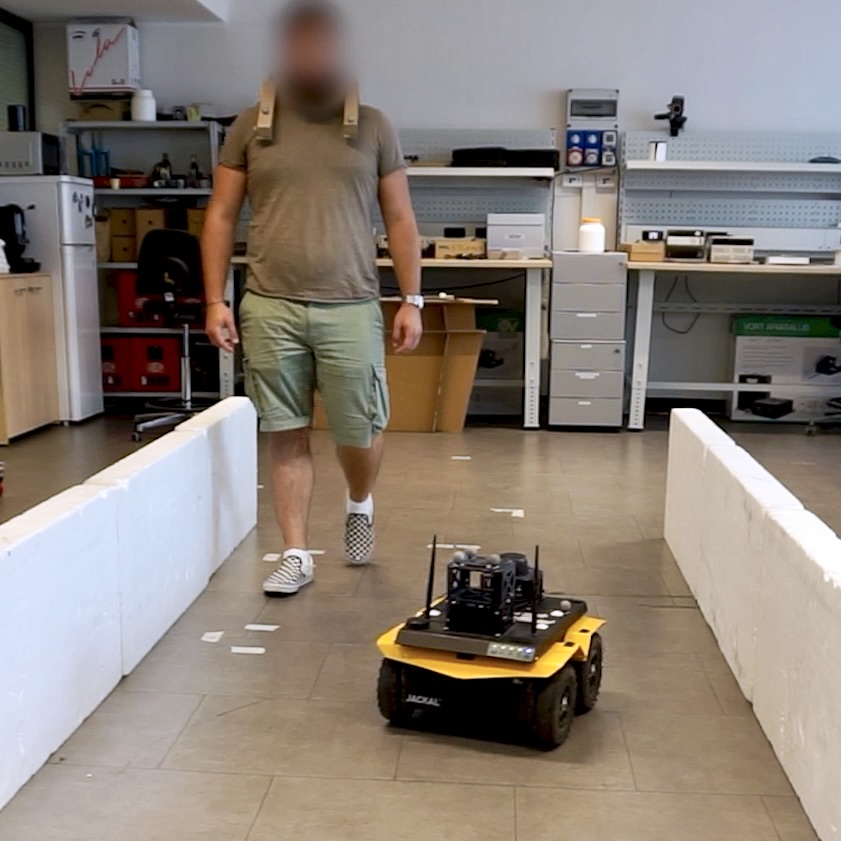}
    }\hfill\subfloat[S2 - Overtaking]{
        \includegraphics[width=0.45\linewidth]{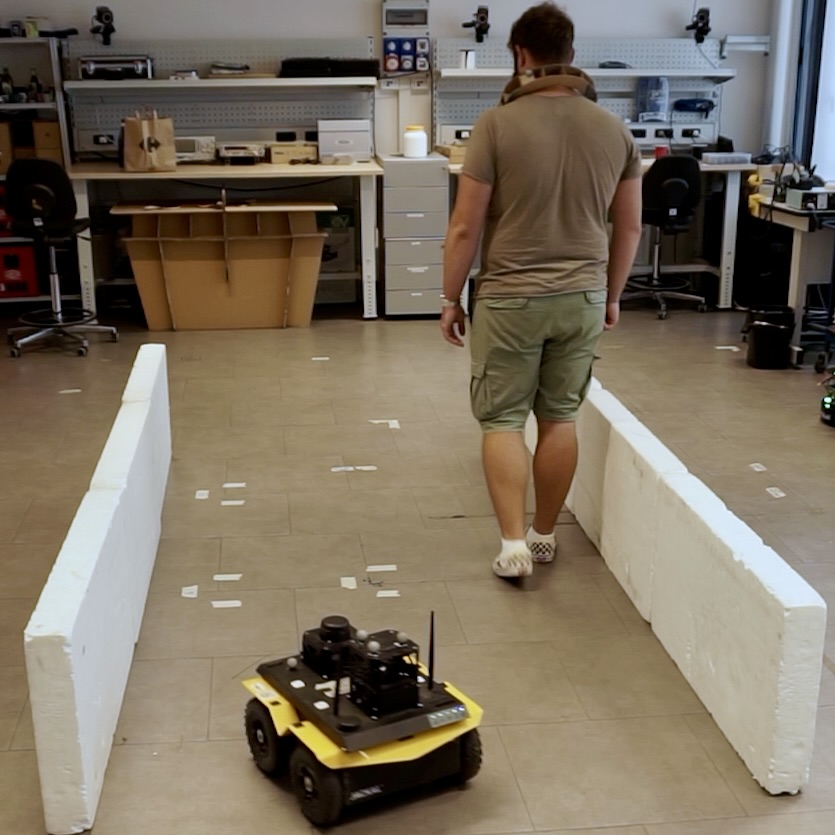}
    }
    \caption{Examples of navigation in scenarios S1 - Passing and S2 - Overtaking, taken in the laboratory}
    \label{fig:experiment_lab}
\end{figure}

\subsection{Scenario Descriptions}
\label{subsec:scenario_descriptions}
\begin{figure*}[!ht]
    \centering
    \subfloat[S1 - Passing]{
       \includegraphics[width=0.22\linewidth]{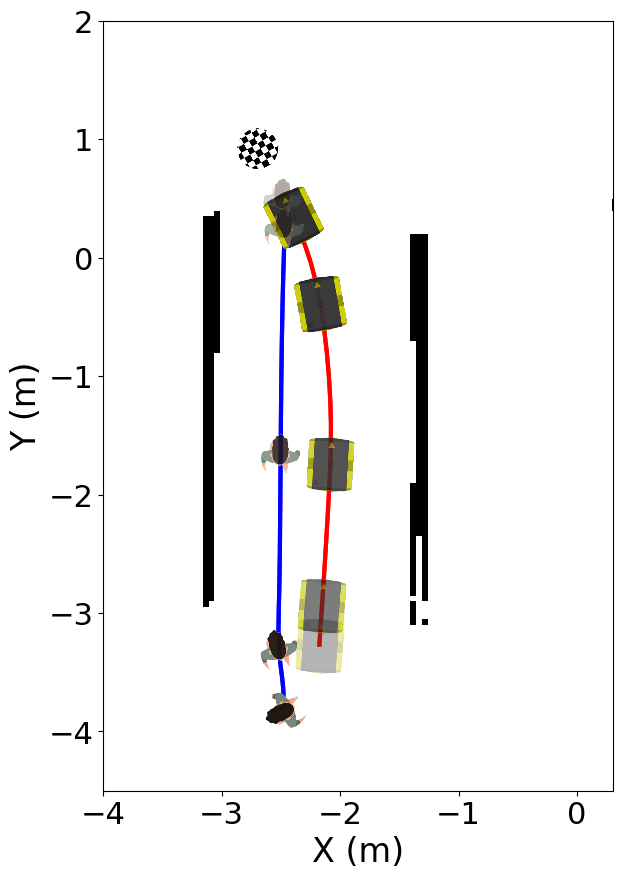}
       \label{fig:passing}
    }
    \subfloat[S2 - Overtaking]{
       \includegraphics[width=0.22\linewidth]{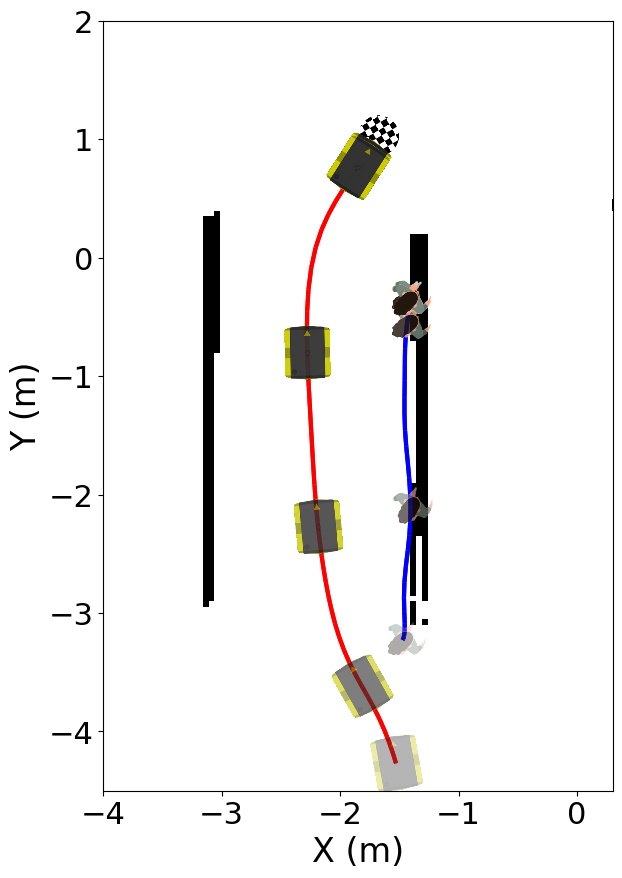}
       \label{fig:overtaking}
    }
    \subfloat[S3 - Crossing 1]{
       \includegraphics[width=0.22\linewidth]{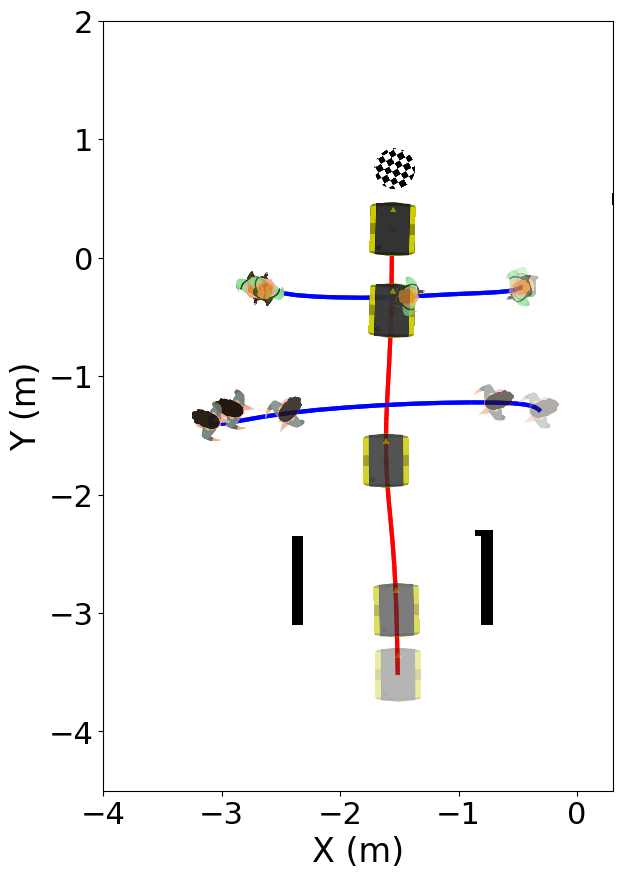}
       \label{fig:crossing_1}
    }
    \subfloat[S4 - Crossing 2]{
       \includegraphics[width=0.22\linewidth]{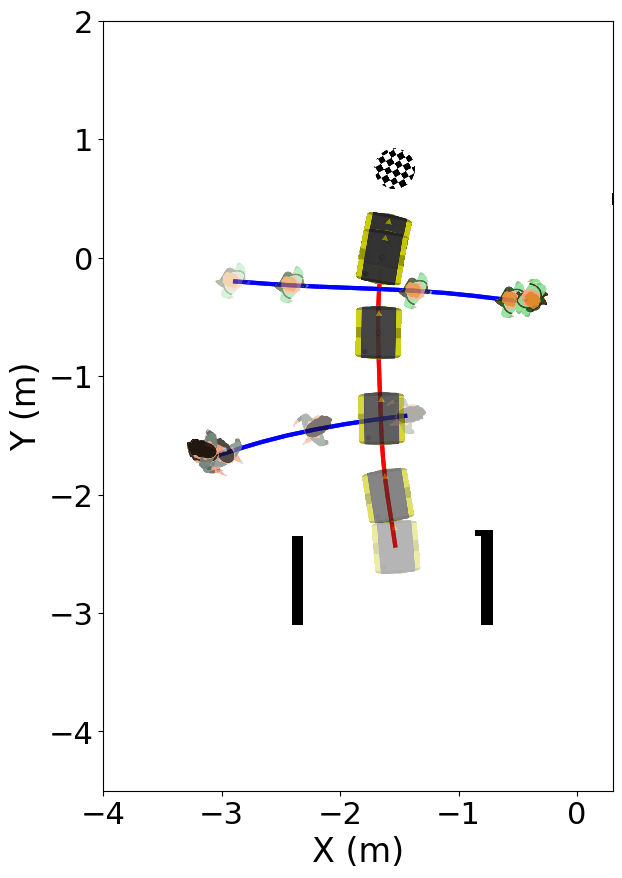}
       \label{fig:crossing_2}
    }
    \\
    \subfloat[S5 - Crossing 3]{
       \includegraphics[width=0.22\linewidth]{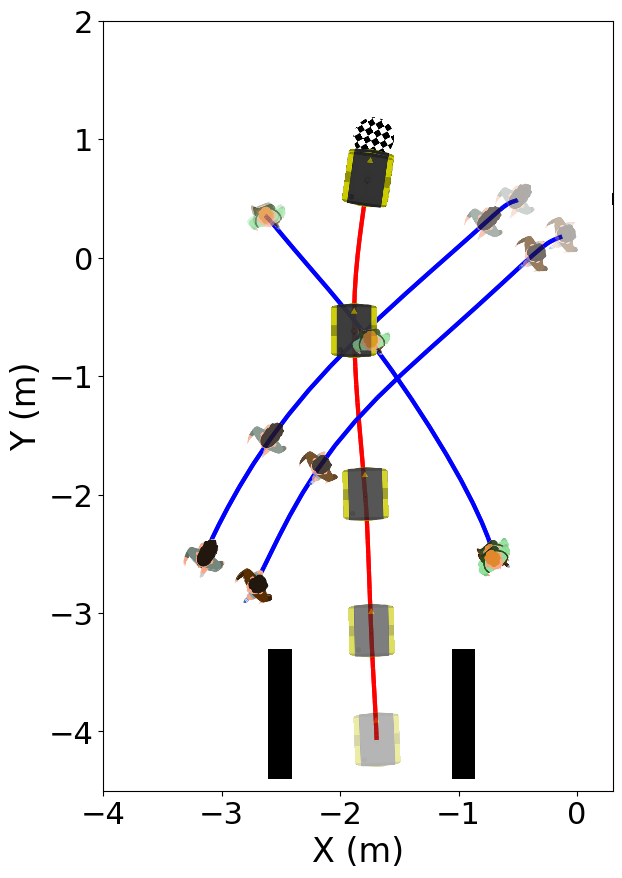}
       \label{fig:crossing_3}
    }
    \subfloat[S6 - Narrow Turn]{
       \includegraphics[width=0.22\linewidth]{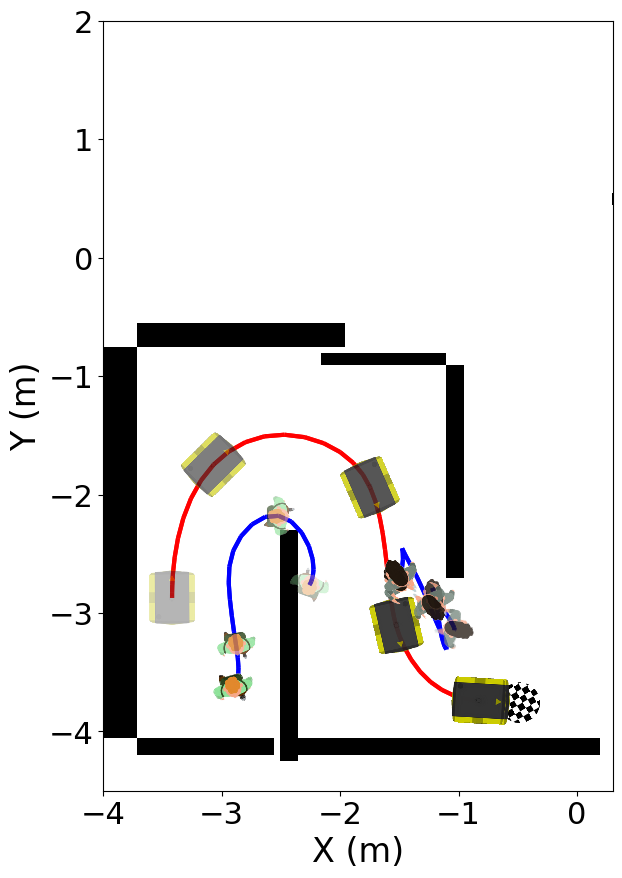}
       \label{fig:narrow_turn}
    }
    \subfloat[S7 - Mixed]{
       \includegraphics[width=0.22\linewidth]{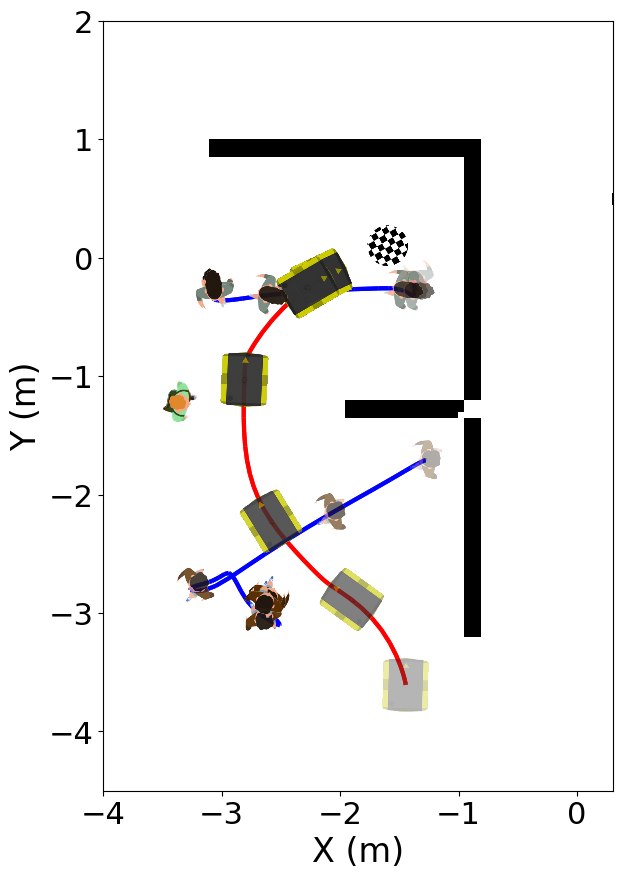}
       \label{fig:mixed}
    }
    \subfloat[S8 - Curious person]{
       \includegraphics[width=0.22\linewidth]{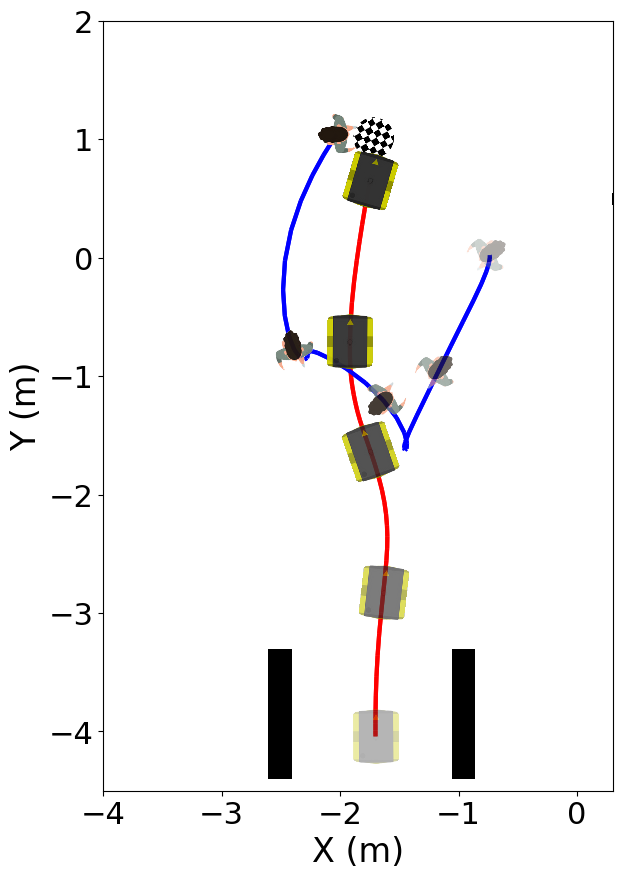}
       \label{fig:curious_person}
    }
    \caption
    {Trajectories associated to a successful run for each of the eight scenarios}
    \label{fig:scenario_maps}
\end{figure*}

The scenarios designed for categorized benchmarking include the most common and representative situations of human motion interaction, as well as more complex, or peculiar settings that may take place in real life. The scenarios are described in Table \ref{tab:scenarios_description}, and the associated maps are shown in Fig. \ref{fig:scenario_maps}, along with a sample of robot and agents trajectories.
These scenarios were also assigned a preliminary difficulty rating based on the number of agents present and the particular environment the robot must traverse. This rating is intended only to be indicative for the reader; it was not involved in the evaluation of the metrics. 
Corridor passing and overtaking situations are tested in scenarios S1 and S2, while different crossing conditions are created in scenarios S3, S4, and S5. Narrow passages and hybrid people trajectories characterize scenarios S6 and S7.
In these first seven scenarios, the humans walk at a moderate speed, between \qtyrange{0.4}{1}{\meter\per\second}, treating the robot as another person and modifying their trajectories only to avoid potential collisions. In contrast, in the scenario S8, a person chases and actively disturbs the robot's motion, acting as a curious person attracted by the robot.
For each scenario, Online Resource 1 shows an example of a favorable and an unfavorable run.

\begin{figure*}[htbp]
    \centering
    \includegraphics[width=\linewidth]{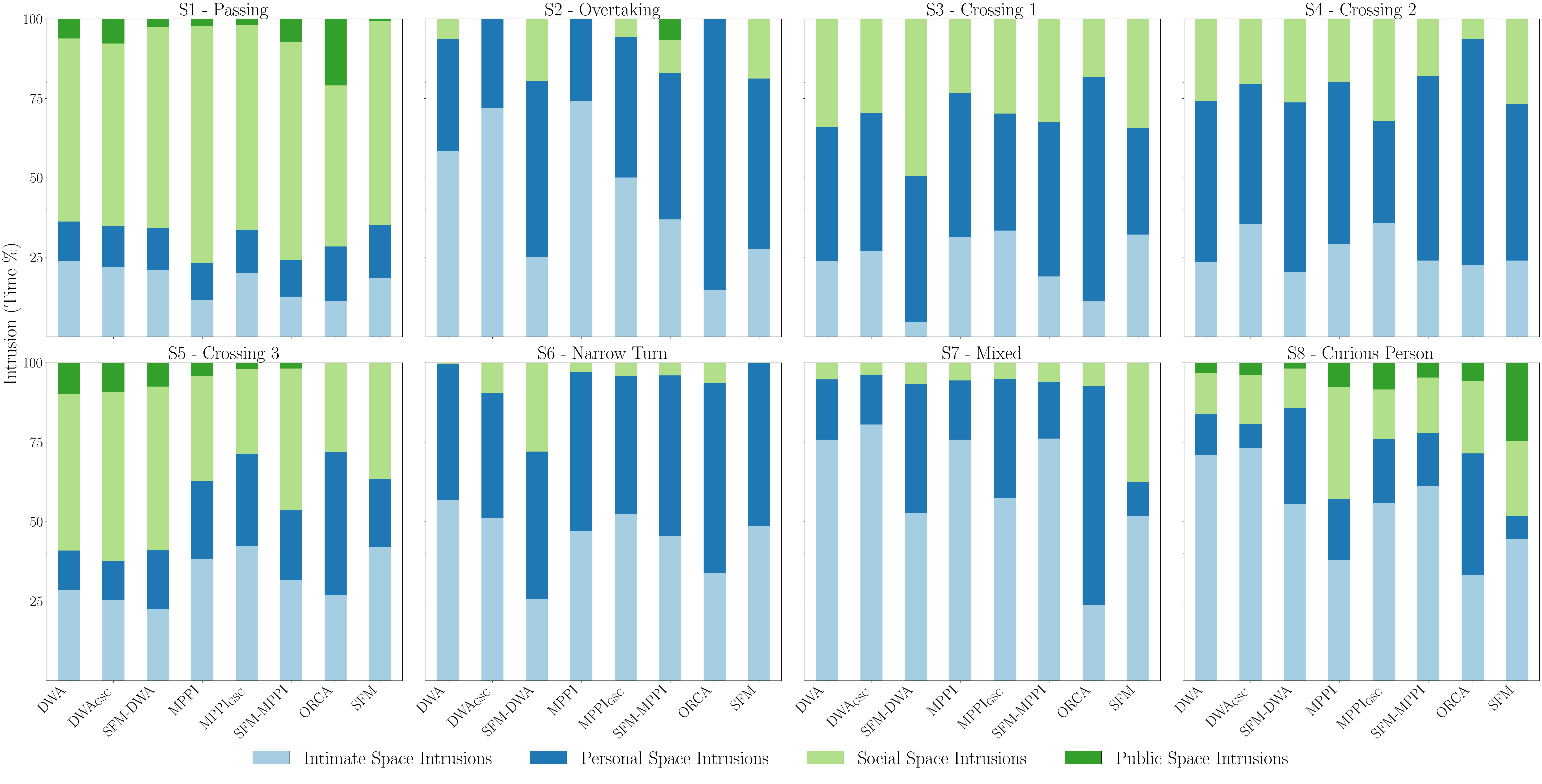}
    \caption{Proxemics metrics as percentage of time spent in social spaces for each scenario.}
    \label{fig:proxemics_metrics}
\end{figure*}

\begin{figure}[t]
    \centering
    \includegraphics[width=\linewidth]{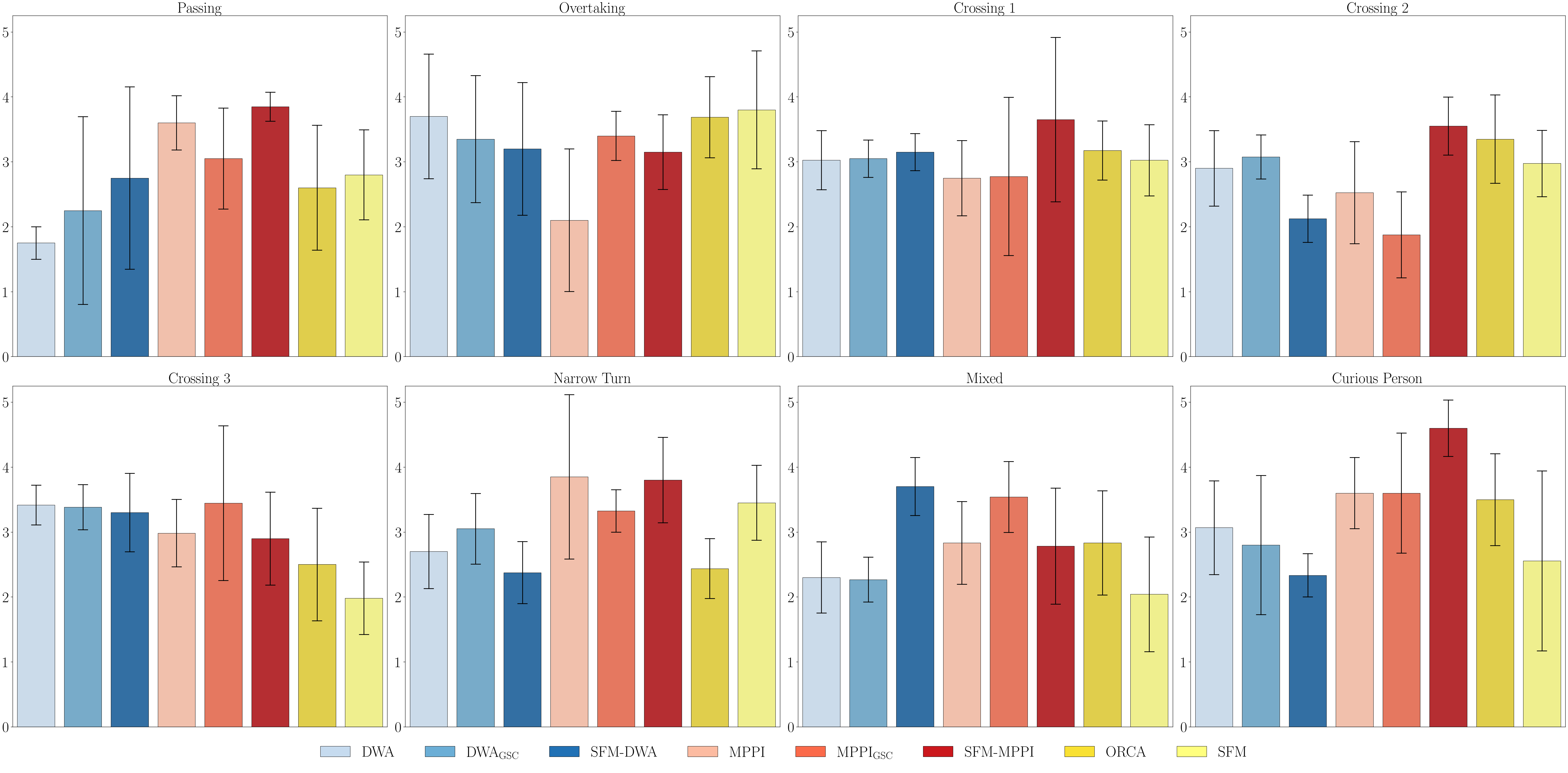}
    \caption{Overall human evaluation presented as bar plots, reporting the mean values and standard deviations for all scenarios. SFM-MPPI tends to rank at the higher end of human preference.}
    \label{fig:human_evaluation}
\end{figure}


\section{Results and Discussion}
\label{sec:results}
This section discusses the quantitative and qualitative results obtained from the experimental benchmarking activity, composed of five runs for each scenario and local planner, totally performing 320 experiments evaluated through metrics and human surveys. These results include 278 successful runs and 42 failures divided in 7 collisions with an agent, 8 collisions with the environment, 24 timeouts, and 3 navigation failures.
Differences between controllers were tested separately for each scenario with Kruskal-Wallis tests, followed by Dunn post-hoc comparisons; controllers with fewer than two valid runs in a scenario were excluded from that test, and $p$-values were adjusted with the Holm method across all tests of the same family (40 tests for the quantitative metrics, 39 for the human ratings). With five runs per combination, the statistical power is limited: after correction, significant differences emerged for CHC in all scenarios except S6, TTG in S1 and S2, PL in S7, and AMD in S5 and S7, while differences in $\mathrm{SW_{s}}$ did not reach significance. The remaining comparisons discussed below should therefore be read as tendencies.
 
\subsection{Quantitative Results}

The quantitative results of the benchmark are reported in Table \ref{tab:quantitative_results}. 
The results show that each scenario poses different challenges and requires different behaviors, making the benchmark ranking highly variable. Thus, categorizing the experiment and its complexity is a relevant procedure to present a significant results analysis in social navigation.
The easiest scenarios proved to be the ones indicated with \textit{Moderate} complexity in Table \ref{tab:scenarios_description}, with an average SR of at least \qty{90}{\percent}, clearly separating them from the \textit{Hard} scenarios.
In this first set of experiments we can already detect clear trends in the performance of the eight controllers: (i) crowd models such as ORCA and SFM do not consistently outperform the local planners in terms of success rate and socially relevant metrics such as average minimum distance to people (AMD) and social work per second $\mathrm{SW_{s}}$, while showing markedly higher heading changes; moreover, ORCA often shows strongly uncertain motion, leading to collision in some passing or crossing runs, and to a generally unnatural heading of the robot; (ii) DWA and MPPI baselines present comparable performance, with DWA showing more stable trajectories and slightly better handling of crossing situations; (iii) the adoption of social cost plugins improves safety and social metrics in both DWA and MPPI. The Gaussian costmap generally provided the highest reliability for both families of controllers,  while the SFM-based cost generally improved the social metrics compared to the other two versions of the same planner, lowering $\mathrm{SW_{s}}$ in seven and raising AMD in five of the eight cases, at the cost of a slight increase in path length and time. 
According to these findings, the SFM-DWA and the SFM-MPPI can be considered reasonable choices for these scenarios.
Crossing 3, Narrow Turn, and Mixed scenarios proved to be the most challenging for all the algorithms, having a lower average SR compared to the rest, with S6 exhibiting only a \qty{67.5}{\percent} average SR. These scenes revealed more subtle differences and limits of the controllers.
Although the previous set of considerations still applies in general, we find that the DWA better handles obstacle avoidance in both people crossing and narrow turns, and is reliable in most cases. Nonetheless, the overly cautious nature of the SFM-DWA introduced uncertainty in narrow, densely populated passages, leading it to fail in four of the five runs. These failures occurred because the robot stopped while turning the corner, just as the pedestrian appeared. The SFM-MPPI, however, succeeded in all runs and kept a higher AMD value than the other MPPI variants (\qty{0.52}{\meter}), while maintaining a low TTG (\qty{16.5}{s}). The other MPPI variants succeeded in only three of the five runs, although the MPPI baseline reached the goal faster. ORCA was strongly affected by static obstacles and narrow passages, whilst SFM alone kept consistent navigation and social metrics despite the drop in success rate.
The Crossing 3 scenario favors the DWA over the MPPI family, with a lower $\mathrm{SW_{s}}$, a higher AMD and 100 \unit{\percent} SR, with the $\mathrm{DWA_{GSC}}$ reaching the highest AMD.
In the Mixed scenario, the DWA family keeps a 100 \unit{\percent} SR and a lower $\mathrm{SW_{s}}$, but AMD is low for most controllers; the significant differences in AMD are between ORCA, which keeps larger distances, and the DWA, $\mathrm{DWA_{GSC}}$, and MPPI. This also shows that the MPPI's proactivity in avoiding obstacles while moving rather than stopping can make it slightly more hazardous in some scenarios. In Crossing 3, the SFM-MPPI also keeps a higher AMD than the simpler MPPI versions.
The Curious Person scenario also shows the reliability of the SFM cost when coupled with the MPPI and its limitations with the DWA, which generally fits better with the Gaussian costmap in complicated scenes.

Overall, the SFM-MPPI resulted in the best trade-off across reliability, social, and navigation metrics. The $\mathrm{DWA_{GSC}}$ was the most reliable implementation, achieving \qty{100}{\percent} SR across all scenarios. The ORCA and SFM often proved to be less reliable and social compliant than expected, with an often uncertain behavior, as can be seen by the high CHC and a low SR in cluttered scenarios. 
The social work plugin favored MPPI over DWA, complementing MPPI's proactive nature and resulting in a favorable balance across the scenarios.

\subsection{Proxemics Results}

The proxemics results are graphically reported in Fig. \ref{fig:proxemics_metrics} as the average percentage of time spent in each social zone. The proxemics metrics show that, overall, ORCA and the SFM-DWA are the most respectful of distances in crowded scenes. ORCA shows the lowest level of intrusion into intimate space, followed by the SFM-DWA, which distinguishes itself from the others, particularly in the crossing scenarios, although both heavily fail, with a \qty{20}{\percent} SR probability in the narrow turn scenario S6. The SFM results in being less effective but generally good at keeping distances at the personal space level. In constrained scenes such as S6 and S7, both DWA and MPPI variants spend a large share of time in the intimate space. Nonetheless, the MPPI-based controllers often rank among the most appreciated in human evaluations, demonstrating that distance preservation is not the only thing that matters to humans. The proxemics histograms show a clear trend with respect to social costs adoption for both DWA and MPPI in the majority of the scenarios, with the social work variant often being the best in minimizing the intimate space intrusion.
 
\begin{table*}[p]
    \centering
    \caption{RESONATE quantitative results for the algorithms described in Table \ref{tab:algorithms_config} along the various scenarios. SR is computed over the five runs of each combination of algorithm and scenario; PL, TTG and CHC are the mean of the successful runs, while $\mathrm{SW_{s}}$ and AMD are the mean of all five runs.
    Best results are highlighted while second best are underlined.
    }
    \footnotesize\setlength{\tabcolsep}{2.5pt}\renewcommand{\arraystretch}{0.85}
    \begin{tabular}{p{1.5cm}lcccccc}
    	\toprule
    	\textbf{Scenario} & \textbf{Navigator} & \textbf{{SR (\unit{\percent})}} $\mathbf{{\uparrow}}$ & \textbf{{PL (\unit{\meter})}} $\mathbf{{\downarrow}}$ & \textbf{{TTG (\unit{\second})}} $\mathbf{{\downarrow}}$ & $\mathbf{{SW_{{s}}}}$ $\mathbf{{\downarrow}}$ & \textbf{{AMD (\unit{\meter})}} $\mathbf{{\uparrow}}$ & \textbf{{CHC (\unit{\radian})}} $\mathbf{{\downarrow}}$ \\
    	\midrule
    	\multirow[t]{8}{=}{\textbf{S1 - Passing}} & DWA & $\mathbf{100}$ & $\mathbf{3.91}$ & $10.99$ & $1.49$ & $1.87$ & $1.09$ \\
    	\textbf{} & $\mathrm{DWA_{GSC}}$ & $\mathbf{100}$ & $4.16$ & $14.60$ & $1.76$ & $1.92$ & $\mathbf{1.02}$ \\
    	\textbf{} & SFM-DWA & $\mathbf{100}$ & $\underline{4.02}$ & $11.55$ & $\mathbf{1.03}$ & $1.88$ & $\underline{1.06}$ \\
    	\textbf{} & MPPI & $\mathbf{100}$ & $4.08$ & $\mathbf{9.19}$ & $1.39$ & $\underline{2.24}$ & $1.25$ \\
    	\textbf{} & $\mathrm{MPPI_{GSC}}$ & $\mathbf{100}$ & $4.22$ & $10.22$ & $1.55$ & $1.93$ & $3.20$ \\
    	\textbf{} & SFM-MPPI & $\mathbf{100}$ & $4.26$ & $10.72$ & $\underline{1.30}$ & $\mathbf{2.30}$ & $2.86$ \\
    	\textbf{} & ORCA & $60$ & $5.71$ & $14.16$ & $4.40$ & $2.22$ & $4.93$ \\
    	\textbf{} & SFM & $\underline{80}$ & $4.06$ & $\underline{9.22}$ & $1.50$ & $1.80$ & $4.81$ \\
    	\midrule
    	\multirow[t]{8}{=}{\textbf{S2 - Overtaking}} & DWA & $\mathbf{100}$ & $5.10$ & $14.33$ & $1.28$ & $0.49$ & $2.68$ \\
    	\textbf{} & $\mathrm{DWA_{GSC}}$ & $\mathbf{100}$ & $4.92$ & $\underline{12.33}$ & $1.52$ & $0.37$ & $\underline{2.13}$ \\
    	\textbf{} & SFM-DWA & $60$ & $5.06$ & $50.40$ & $\mathbf{0.96}$ & $0.82$ & $8.98$ \\
    	\textbf{} & MPPI & $\underline{80}$ & $\underline{4.88}$ & $\mathbf{9.67}$ & $2.50$ & $0.23$ & $\mathbf{1.51}$ \\
    	\textbf{} & $\mathrm{MPPI_{GSC}}$ & $\mathbf{100}$ & $6.13$ & $14.18$ & $1.17$ & $0.50$ & $6.97$ \\
    	\textbf{} & SFM-MPPI & $\underline{80}$ & $5.34$ & $13.85$ & $1.20$ & $\mathbf{0.93}$ & $6.65$ \\
    	\textbf{} & ORCA & $\mathbf{100}$ & $7.02$ & $21.29$ & $1.58$ & $0.61$ & $31.76$ \\
    	\textbf{} & SFM & $\mathbf{100}$ & $\mathbf{4.78}$ & $13.97$ & $\underline{1.09}$ & $\underline{0.92}$ & $24.47$ \\
    	\midrule
    	\multirow[t]{8}{=}{\textbf{S3 - Crossing 1}} & DWA & $\mathbf{100}$ & $\underline{3.95}$ & $11.95$ & $1.01$ & $0.99$ & $\mathbf{1.52}$ \\
    	\textbf{} & $\mathrm{DWA_{GSC}}$ & $\mathbf{100}$ & $\mathbf{3.90}$ & $11.81$ & $1.05$ & $0.90$ & $\underline{1.66}$ \\
    	\textbf{} & SFM-DWA & $\mathbf{100}$ & $4.02$ & $22.63$ & $\mathbf{0.89}$ & $\mathbf{1.20}$ & $3.37$ \\
    	\textbf{} & MPPI & $\mathbf{100}$ & $4.24$ & $\mathbf{10.69}$ & $1.41$ & $0.80$ & $3.63$ \\
    	\textbf{} & $\mathrm{MPPI_{GSC}}$ & $\underline{80}$ & $4.71$ & $13.83$ & $1.82$ & $0.91$ & $4.39$ \\
    	\textbf{} & SFM-MPPI & $\mathbf{100}$ & $4.36$ & $11.39$ & $\underline{0.96}$ & $\underline{1.00}$ & $3.78$ \\
    	\textbf{} & ORCA & $\underline{80}$ & $4.60$ & $14.74$ & $1.26$ & $0.85$ & $13.40$ \\
    	\textbf{} & SFM & $\mathbf{100}$ & $4.93$ & $\underline{11.34}$ & $1.17$ & $0.93$ & $7.06$ \\
    	\midrule
    	\multirow[t]{8}{=}{\textbf{S4 - Crossing 2}} & DWA & $\underline{80}$ & $\mathbf{3.66}$ & $12.15$ & $0.80$ & $\underline{0.92}$ & $\mathbf{1.74}$ \\
    	\textbf{} & $\mathrm{DWA_{GSC}}$ & $\mathbf{100}$ & $\underline{3.78}$ & $\mathbf{11.24}$ & $0.93$ & $0.76$ & $\underline{2.11}$ \\
    	\textbf{} & SFM-DWA & $\underline{80}$ & $4.20$ & $28.70$ & $\underline{0.76}$ & $0.81$ & $4.63$ \\
    	\textbf{} & MPPI & $\mathbf{100}$ & $4.83$ & $14.88$ & $1.16$ & $0.83$ & $13.31$ \\
    	\textbf{} & $\mathrm{MPPI_{GSC}}$ & $\underline{80}$ & $5.23$ & $15.30$ & $1.15$ & $\mathbf{0.94}$ & $6.89$ \\
    	\textbf{} & SFM-MPPI & $\mathbf{100}$ & $4.43$ & $12.90$ & $0.90$ & $0.83$ & $4.88$ \\
    	\textbf{} & ORCA & $\mathbf{100}$ & $6.87$ & $25.09$ & $\mathbf{0.75}$ & $0.64$ & $41.12$ \\
    	\textbf{} & SFM & $\mathbf{100}$ & $4.36$ & $\underline{12.11}$ & $1.07$ & $0.89$ & $10.67$ \\
    	\midrule
    	\multirow[t]{8}{=}{\textbf{S5 - Crossing 3}} & DWA & $\mathbf{100}$ & $\mathbf{4.89}$ & $\mathbf{14.78}$ & $1.07$ & $1.68$ & $\underline{2.87}$ \\
    	\textbf{} & $\mathrm{DWA_{GSC}}$ & $\mathbf{100}$ & $\underline{4.89}$ & $16.64$ & $\underline{1.00}$ & $\mathbf{1.72}$ & $\mathbf{2.72}$ \\
    	\textbf{} & SFM-DWA & $\mathbf{100}$ & $4.98$ & $16.41$ & $\mathbf{0.97}$ & $\underline{1.70}$ & $3.32$ \\
    	\textbf{} & MPPI & $\underline{80}$ & $5.17$ & $16.30$ & $1.13$ & $1.15$ & $10.04$ \\
    	\textbf{} & $\mathrm{MPPI_{GSC}}$ & $40$ & $5.13$ & $\underline{15.03}$ & $4.42$ & $1.01$ & $3.87$ \\
    	\textbf{} & SFM-MPPI & $\underline{80}$ & $5.20$ & $16.33$ & $1.22$ & $1.36$ & $6.06$ \\
    	\textbf{} & ORCA & $\underline{80}$ & $8.64$ & $25.57$ & $1.64$ & $1.02$ & $22.52$ \\
    	\textbf{} & SFM & $\underline{80}$ & $6.43$ & $17.17$ & $1.45$ & $1.11$ & $23.49$ \\
    	\midrule
    	\multirow[t]{8}{=}{\textbf{S6 - Narrow Turn}} & DWA & $\mathbf{100}$ & $6.79$ & $19.86$ & $4.84$ & $0.36$ & $12.44$ \\
    	\textbf{} & $\mathrm{DWA_{GSC}}$ & $\mathbf{100}$ & $\mathbf{5.81}$ & $20.18$ & $4.82$ & $0.50$ & $\mathbf{8.14}$ \\
    	\textbf{} & SFM-DWA & $20$ & $6.44$ & $54.90$ & $\mathbf{3.04}$ & $\mathbf{0.83}$ & $11.17$ \\
    	\textbf{} & MPPI & $60$ & $6.60$ & $\underline{15.26}$ & $5.99$ & $0.50$ & $8.44$ \\
    	\textbf{} & $\mathrm{MPPI_{GSC}}$ & $60$ & $6.81$ & $19.91$ & $4.57$ & $0.47$ & $14.79$ \\
    	\textbf{} & SFM-MPPI & $\mathbf{100}$ & $\underline{6.21}$ & $16.52$ & $\underline{3.75}$ & $0.52$ & $9.34$ \\
    	\textbf{} & ORCA & $20$ & $10.19$ & $31.88$ & $10.68$ & $\underline{0.60}$ & $41.63$ \\
    	\textbf{} & SFM & $\underline{80}$ & $6.28$ & $\mathbf{14.05}$ & $6.48$ & $0.45$ & $\underline{8.14}$ \\
    	\midrule
    	\multirow[t]{8}{=}{\textbf{S7 - Mixed}} & DWA & $\mathbf{100}$ & $\underline{4.85}$ & $28.71$ & $1.82$ & $0.25$ & $\underline{6.72}$ \\
    	\textbf{} & $\mathrm{DWA_{GSC}}$ & $\mathbf{100}$ & $\mathbf{4.64}$ & $20.86$ & $1.50$ & $0.23$ & $\mathbf{5.01}$ \\
    	\textbf{} & SFM-DWA & $\mathbf{100}$ & $7.13$ & $22.11$ & $\underline{1.30}$ & $0.53$ & $15.42$ \\
    	\textbf{} & MPPI & $60$ & $6.25$ & $\underline{17.87}$ & $2.11$ & $0.27$ & $15.81$ \\
    	\textbf{} & $\mathrm{MPPI_{GSC}}$ & $\underline{80}$ & $9.92$ & $31.95$ & $4.31$ & $0.45$ & $29.12$ \\
    	\textbf{} & SFM-MPPI & $\underline{80}$ & $6.18$ & $22.88$ & $2.32$ & $0.32$ & $15.64$ \\
    	\textbf{} & ORCA & $\mathbf{100}$ & $6.39$ & $24.70$ & $2.46$ & $\underline{0.69}$ & $28.23$ \\
    	\textbf{} & SFM & $40$ & $6.31$ & $\mathbf{16.23}$ & $\mathbf{1.27}$ & $\mathbf{0.78}$ & $19.30$ \\
    	\midrule
    	\multirow[t]{8}{=}{\textbf{S8 - Curious Person}} & DWA & $\mathbf{100}$ & $\underline{5.42}$ & $24.15$ & $1.12$ & $0.57$ & $\underline{6.74}$ \\
    	\textbf{} & $\mathrm{DWA_{GSC}}$ & $\mathbf{100}$ & $\mathbf{4.96}$ & $21.45$ & $1.13$ & $0.62$ & $\mathbf{4.06}$ \\
    	\textbf{} & SFM-DWA & $\underline{80}$ & $7.70$ & $33.60$ & $0.98$ & $0.69$ & $22.59$ \\
    	\textbf{} & MPPI & $\mathbf{100}$ & $7.79$ & $\underline{18.88}$ & $\mathbf{0.75}$ & $\underline{1.17}$ & $15.02$ \\
    	\textbf{} & $\mathrm{MPPI_{GSC}}$ & $\mathbf{100}$ & $10.40$ & $\mathbf{17.94}$ & $0.94$ & $0.96$ & $21.19$ \\
    	\textbf{} & SFM-MPPI & $\mathbf{100}$ & $6.67$ & $19.24$ & $\underline{0.87}$ & $0.84$ & $13.84$ \\
    	\textbf{} & ORCA & $\mathbf{100}$ & $10.90$ & $36.08$ & $1.30$ & $1.14$ & $60.98$ \\
    	\textbf{} & SFM & $60$ & $6.62$ & $23.70$ & $1.32$ & $\mathbf{1.53}$ & $69.48$ \\
    	\bottomrule
    \end{tabular}
    \label{tab:quantitative_results}
\end{table*}

\begin{table*}[p]
    \centering
    \caption{RESONATE qualitative results for the algorithms described in Table \ref{tab:algorithms_config} along the various scenarios. The results are the mean of all rated runs, including the unsuccessful ones, for each combination of algorithm and scenario.
    Best results are highlighted while second best are underlined.
    }
    \footnotesize\setlength{\tabcolsep}{4pt}\renewcommand{\arraystretch}{0.85}
    \begin{tabular}{p{1.5cm}lccccc}
    	\toprule
    	\textbf{Scenario} & \textbf{Navigator} & \textbf{{SA}} $\mathbf{{\uparrow}}$ & \textbf{{FR}} $\mathbf{{\uparrow}}$ & $\mathbf{{NA}}$ $\mathbf{{\uparrow}}$ & \textbf{{AF}} $\mathbf{{\uparrow}}$ & \textbf{{Overall}} $\mathbf{{\uparrow}}$ \\
    	\midrule
    	\multirow[t]{8}{=}{\textbf{S1 - Passing}} & DWA & $1.40$ & $1.40$ & $2.40$ & $1.80$ & $1.75$ \\
    	\textbf{} & $\mathrm{DWA_{GSC}}$ & $2.40$ & $2.00$ & $2.40$ & $2.20$ & $2.25$ \\
    	\textbf{} & SFM-DWA & $2.80$ & $2.20$ & $\underline{3.20}$ & $2.80$ & $2.75$ \\
    	\textbf{} & MPPI & $\underline{3.60}$ & $\underline{3.20}$ & $\mathbf{3.40}$ & $\mathbf{4.20}$ & $\underline{3.60}$ \\
    	\textbf{} & $\mathrm{MPPI_{GSC}}$ & $3.40$ & $3.00$ & $2.60$ & $\underline{3.20}$ & $3.05$ \\
    	\textbf{} & SFM-MPPI & $\mathbf{4.00}$ & $\mathbf{3.80}$ & $\mathbf{3.40}$ & $\mathbf{4.20}$ & $\mathbf{3.85}$ \\
    	\textbf{} & ORCA & $3.20$ & $2.80$ & $1.60$ & $2.80$ & $2.60$ \\
    	\textbf{} & SFM & $3.00$ & $2.60$ & $2.60$ & $3.00$ & $2.80$ \\
    	\midrule
    	\multirow[t]{8}{=}{\textbf{S2 - Overtaking}} & DWA & $\underline{3.80}$ & $\underline{3.60}$ & $\mathbf{3.80}$ & $3.60$ & $\underline{3.70}$ \\
    	\textbf{} & $\mathrm{DWA_{GSC}}$ & $3.40$ & $3.40$ & $3.40$ & $3.20$ & $3.35$ \\
    	\textbf{} & SFM-DWA & $\mathbf{4.00}$ & $3.20$ & $1.40$ & $\underline{4.20}$ & $3.20$ \\
    	\textbf{} & MPPI & $1.80$ & $1.80$ & $\underline{3.60}$ & $1.20$ & $2.10$ \\
    	\textbf{} & $\mathrm{MPPI_{GSC}}$ & $\underline{3.80}$ & $3.40$ & $3.00$ & $3.40$ & $3.40$ \\
    	\textbf{} & SFM-MPPI & $3.40$ & $3.40$ & $2.60$ & $3.20$ & $3.15$ \\
    	\textbf{} & ORCA & $\mathbf{4.00}$ & $3.25$ & $3.25$ & $\mathbf{4.25}$ & $3.69$ \\
    	\textbf{} & SFM & $\underline{3.80}$ & $\mathbf{3.80}$ & $3.40$ & $\underline{4.20}$ & $\mathbf{3.80}$ \\
    	\midrule
    	\multirow[t]{8}{=}{\textbf{S3 - Crossing 1}} & DWA & $3.10$ & $3.00$ & $\underline{2.90}$ & $3.10$ & $3.02$ \\
    	\textbf{} & $\mathrm{DWA_{GSC}}$ & $3.30$ & $3.00$ & $2.70$ & $3.20$ & $3.05$ \\
    	\textbf{} & SFM-DWA & $\mathbf{4.10}$ & $3.00$ & $1.30$ & $\mathbf{4.20}$ & $3.15$ \\
    	\textbf{} & MPPI & $2.50$ & $2.50$ & $\underline{2.90}$ & $3.10$ & $2.75$ \\
    	\textbf{} & $\mathrm{MPPI_{GSC}}$ & $3.00$ & $2.80$ & $2.60$ & $2.70$ & $2.77$ \\
    	\textbf{} & SFM-MPPI & $\underline{3.70}$ & $\mathbf{3.70}$ & $\mathbf{3.70}$ & $3.50$ & $\mathbf{3.65}$ \\
    	\textbf{} & ORCA & $3.10$ & $\underline{3.50}$ & $2.50$ & $\underline{3.60}$ & $\underline{3.17}$ \\
    	\textbf{} & SFM & $3.40$ & $3.10$ & $2.60$ & $3.00$ & $3.02$ \\
    	\midrule
    	\multirow[t]{8}{=}{\textbf{S4 - Crossing 2}} & DWA & $3.00$ & $2.60$ & $\underline{3.00}$ & $3.00$ & $2.90$ \\
    	\textbf{} & $\mathrm{DWA_{GSC}}$ & $3.20$ & $3.00$ & $2.90$ & $3.20$ & $3.08$ \\
    	\textbf{} & SFM-DWA & $2.50$ & $2.00$ & $1.50$ & $2.50$ & $2.12$ \\
    	\textbf{} & MPPI & $2.90$ & $2.30$ & $2.30$ & $2.60$ & $2.52$ \\
    	\textbf{} & $\mathrm{MPPI_{GSC}}$ & $1.90$ & $1.90$ & $1.80$ & $1.90$ & $1.88$ \\
    	\textbf{} & SFM-MPPI & $\mathbf{3.90}$ & $\mathbf{3.40}$ & $\mathbf{3.20}$ & $\underline{3.70}$ & $\mathbf{3.55}$ \\
    	\textbf{} & ORCA & $\underline{3.70}$ & $\underline{3.30}$ & $2.60$ & $\mathbf{3.80}$ & $\underline{3.35}$ \\
    	\textbf{} & SFM & $3.30$ & $2.90$ & $2.50$ & $3.20$ & $2.98$ \\
    	\midrule
    	\multirow[t]{8}{=}{\textbf{S5 - Crossing 3}} & DWA & $\underline{3.60}$ & $\underline{3.47}$ & $3.07$ & $\underline{3.53}$ & $\underline{3.42}$ \\
    	\textbf{} & $\mathrm{DWA_{GSC}}$ & $\mathbf{3.80}$ & $\mathbf{3.67}$ & $2.73$ & $3.33$ & $3.38$ \\
    	\textbf{} & SFM-DWA & $3.53$ & $3.27$ & $3.00$ & $3.40$ & $3.30$ \\
    	\textbf{} & MPPI & $2.87$ & $3.07$ & $3.07$ & $2.93$ & $2.98$ \\
    	\textbf{} & $\mathrm{MPPI_{GSC}}$ & $3.44$ & $\mathbf{3.67}$ & $\underline{3.11}$ & $\mathbf{3.56}$ & $\mathbf{3.44}$ \\
    	\textbf{} & SFM-MPPI & $2.80$ & $2.93$ & $\mathbf{3.20}$ & $2.67$ & $2.90$ \\
    	\textbf{} & ORCA & $3.28$ & $2.00$ & $1.33$ & $3.39$ & $2.50$ \\
    	\textbf{} & SFM & $2.17$ & $1.83$ & $2.00$ & $1.92$ & $1.98$ \\
    	\midrule
    	\multirow[t]{8}{=}{\textbf{S6 - Narrow Turn}} & DWA & $2.60$ & $2.30$ & $3.10$ & $2.80$ & $2.70$ \\
    	\textbf{} & $\mathrm{DWA_{GSC}}$ & $3.20$ & $3.10$ & $2.40$ & $3.50$ & $3.05$ \\
    	\textbf{} & SFM-DWA & $3.62$ & $2.25$ & $1.00$ & $2.62$ & $2.38$ \\
    	\textbf{} & MPPI & $\mathbf{4.00}$ & $\underline{3.60}$ & $\mathbf{3.90}$ & $\mathbf{3.90}$ & $\mathbf{3.85}$ \\
    	\textbf{} & $\mathrm{MPPI_{GSC}}$ & $3.20$ & $3.30$ & $3.30$ & $3.50$ & $3.33$ \\
    	\textbf{} & SFM-MPPI & $\underline{3.70}$ & $\mathbf{3.70}$ & $\mathbf{3.90}$ & $\mathbf{3.90}$ & $\underline{3.80}$ \\
    	\textbf{} & ORCA & $3.38$ & $2.25$ & $1.75$ & $2.38$ & $2.44$ \\
    	\textbf{} & SFM & $3.40$ & $3.30$ & $\underline{3.50}$ & $\underline{3.60}$ & $3.45$ \\
    	\midrule
    	\multirow[t]{8}{=}{\textbf{S7 - Mixed}} & DWA & $2.67$ & $2.20$ & $1.87$ & $2.47$ & $2.30$ \\
    	\textbf{} & $\mathrm{DWA_{GSC}}$ & $2.47$ & $2.07$ & $2.27$ & $2.27$ & $2.27$ \\
    	\textbf{} & SFM-DWA & $\mathbf{4.20}$ & $\mathbf{3.80}$ & $\underline{3.07}$ & $\mathbf{3.73}$ & $\mathbf{3.70}$ \\
    	\textbf{} & MPPI & $2.80$ & $2.67$ & $2.93$ & $2.93$ & $2.83$ \\
    	\textbf{} & $\mathrm{MPPI_{GSC}}$ & $\underline{3.75}$ & $\underline{3.58}$ & $\mathbf{3.42}$ & $\underline{3.42}$ & $\underline{3.54}$ \\
    	\textbf{} & SFM-MPPI & $2.73$ & $2.67$ & $2.87$ & $2.87$ & $2.78$ \\
    	\textbf{} & ORCA & $3.22$ & $2.78$ & $2.11$ & $3.22$ & $2.83$ \\
    	\textbf{} & SFM & $2.00$ & $2.00$ & $2.33$ & $1.83$ & $2.04$ \\
    	\midrule
    	\multirow[t]{8}{=}{\textbf{S8 - Curious Person}} & DWA & - & $3.40$ & $2.80$ & $3.00$ & $3.07$ \\
    	\textbf{} & $\mathrm{DWA_{GSC}}$ & - & $2.80$ & $2.80$ & $2.80$ & $2.80$ \\
    	\textbf{} & SFM-DWA & - & $2.20$ & $2.40$ & $2.40$ & $2.33$ \\
    	\textbf{} & MPPI & - & $\underline{3.60}$ & $\underline{3.80}$ & $3.40$ & $\underline{3.60}$ \\
    	\textbf{} & $\mathrm{MPPI_{GSC}}$ & - & $\underline{3.60}$ & $\underline{3.80}$ & $3.40$ & $\underline{3.60}$ \\
    	\textbf{} & SFM-MPPI & - & $\mathbf{4.60}$ & $\mathbf{4.40}$ & $\mathbf{4.80}$ & $\mathbf{4.60}$ \\
    	\textbf{} & ORCA & - & $3.50$ & $2.50$ & $\underline{4.50}$ & $3.50$ \\
    	\textbf{} & SFM & - & $2.67$ & $2.00$ & $3.00$ & $2.56$ \\
    	\bottomrule
    \end{tabular}
    \label{tab:qualitative_results}
\end{table*}

\subsection{Human Evaluation}
The qualitative results obtained from participants' surveys are shown in Table \ref{tab:qualitative_results}. The safety metric was not evaluated for scenario S8 - \textit{Curious Person} because the person was actively seeking close interaction with the robot, which interfered with path planning, rendering the metric meaningless in this instance. An overall score was computed as the mean of all human evaluation metrics (for S8 the mean over the 3 metrics available), and the results were reported as bar plots showing the mean values and standard deviations for each scenario in Fig. \ref{fig:human_evaluation}. 
The SFM-MPPI consistently scores towards the higher end of human ratings, especially reaching the top in the \textit{Passing}, \textit{Crossing 1}, \textit{Crossing 2}, and \textit{Curious Person} (and second-highest in \textit{Narrow Turn}) scenarios, where its predictive planning over the future horizon allowed it to navigate effectively also in constrained spaces and led to improved social acceptance. In terms of specific metrics, SFM-MPPI often achieves the best ratings in friendliness, naturalness, and avoidance foresight, highlighting its fluency in anticipating critical situations and acting in accordance with social norms. Also, the unpredictable behavior of a curious person is better handled by the MPPI family of controllers. On the other hand, human evaluations confirmed the findings of the proxemics study, identifying the DWA family and the SFM-DWA as often the safest. The \textit{Mixed} scenario particularly favors the SFM-DWA, although resulting in a generally unnatural trajectory, looking at the naturalness score.

In general, coupling the SFM-based social cost with DWA and MPPI controllers results in a winning strategy to obtain both reliability and social acceptance. The same is not true for the adoption of \textit{Gaussian costmaps}, in the case of DWA we have mostly identical rating with $\mathrm{DWA_{GSC}}$, in the MPPI case the results are instead unstable, the tendencies may either drop (e.g. \textit{Crossing 2})or rise (e.g. \textit{Overtaking} and \textit{Mixed}).
The Overtaking scenario favored the SFM, with the ORCA also performing well. Nonetheless, these crowd-inspired models did not excel in many cases, particularly in naturalness, due to their uncertain and oscillatory behavior.
Among the differences in human ratings, only naturalness in the Narrow Turn scenario was significant after correction, with the SFM-DWA rated lower than the MPPI and the SFM-MPPI; the other rankings should be read as tendencies.



\subsection{Cross-metrics analysis}
\begin{figure}[t]
    \centering
    \subfloat[S1 - Passing]{
       \label{fig:passing_cross}
       \includegraphics[width=0.98\linewidth]{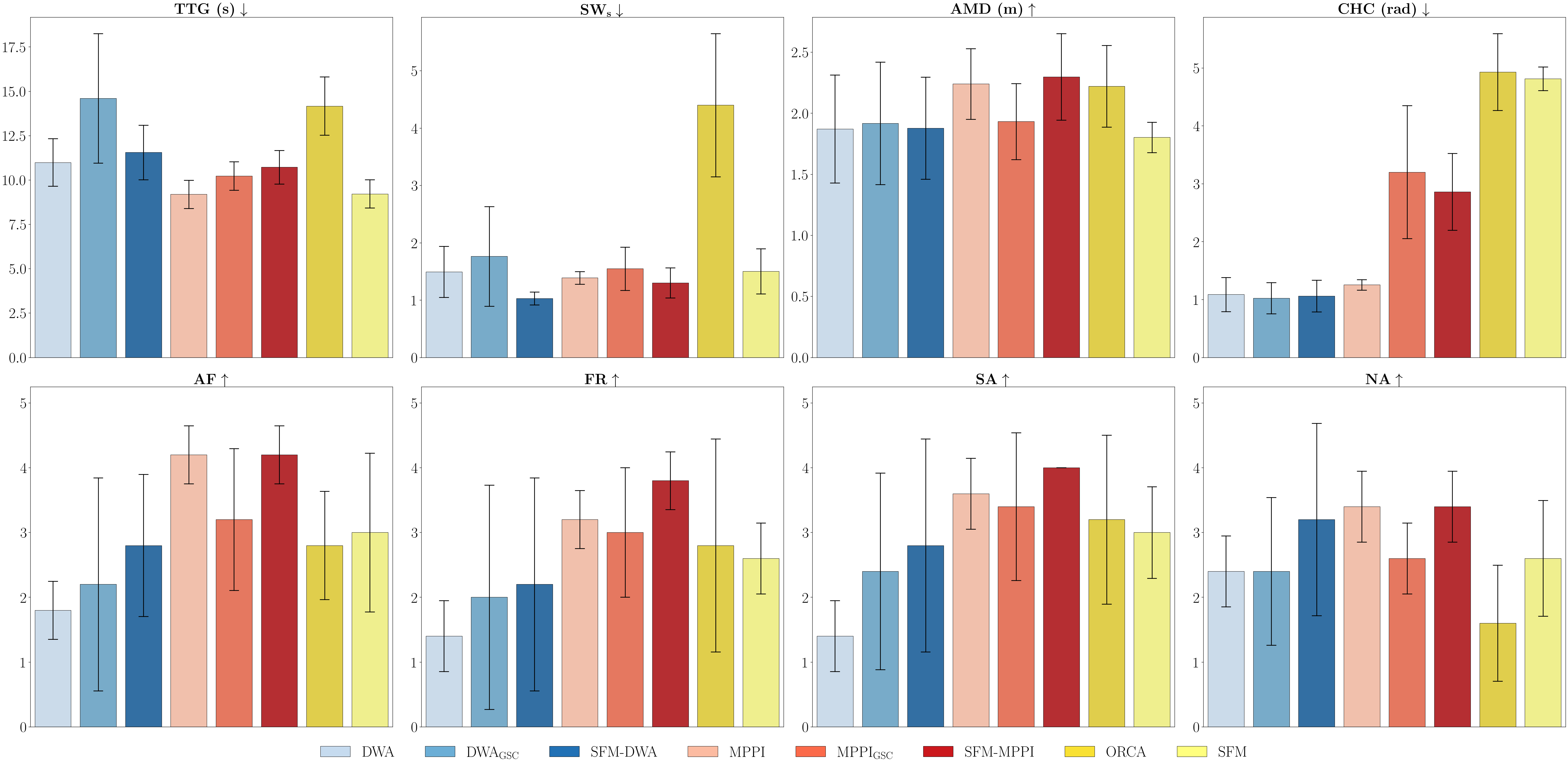}
    } \\
    \subfloat[S3 - Crossing 1]{
       \label{fig:crossing_cross}
       \includegraphics[width=0.98\linewidth]{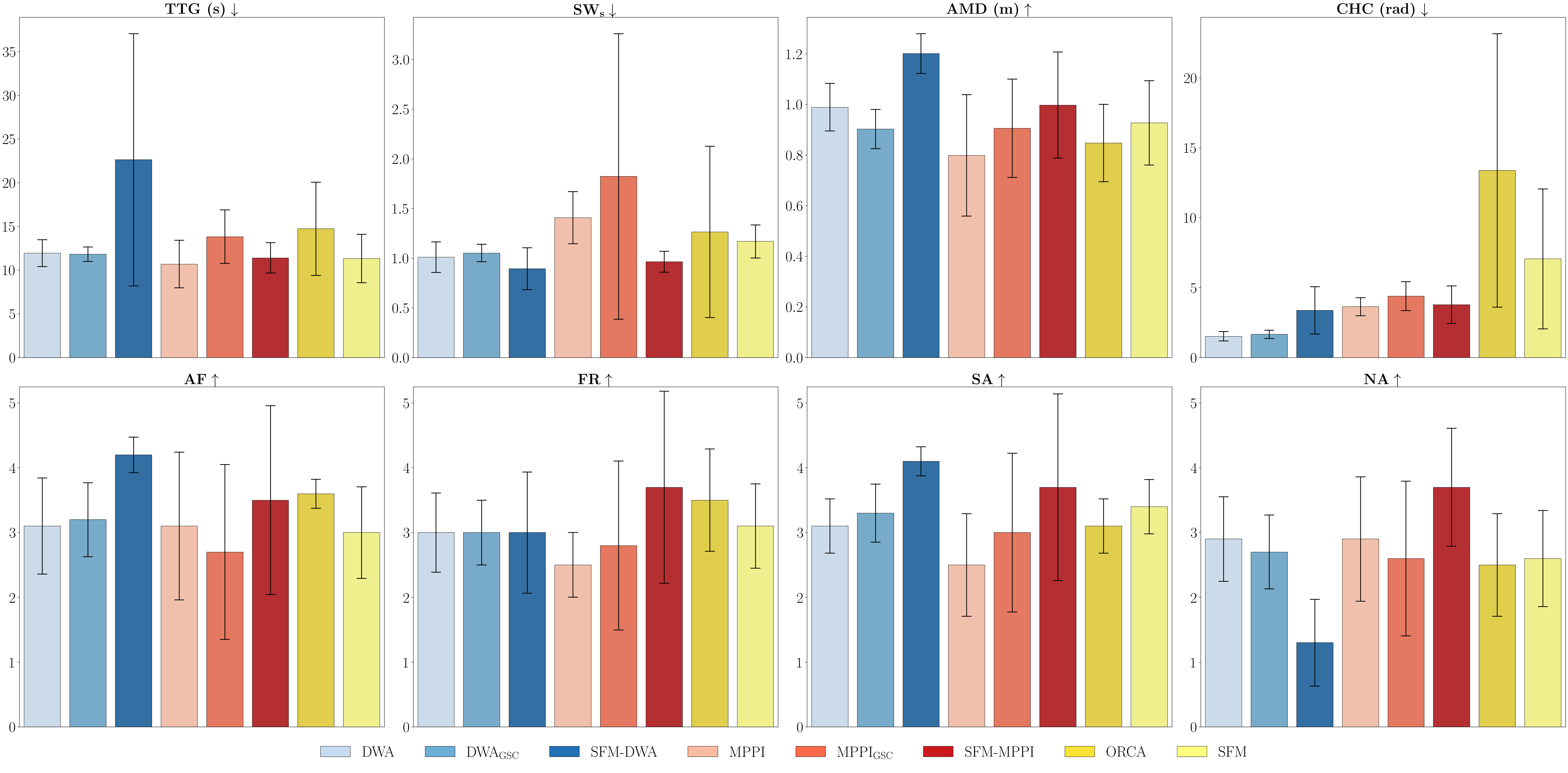}
    }
    \caption{Cross-metrics analysis. The figure shows that some correlations between metrics can be observed, as is the example of the AMD showing a similar trend to the Safety metric. Fig. \ref{fig:passing_cross} shows the results for the Passing scenario, while Fig. \ref{fig:crossing_cross} shows the result for the Crossing 1 scenario}
    \label{fig:passing_and_crossing_cross_analysis}
\end{figure}

The previous analysis delved into a general comparison across all the scenarios separating the quantitative and qualitative characteristics. The results of our benchmark, however, also allow to inspect possible correlations between the two families of results. Fig. \ref{fig:passing_and_crossing_cross_analysis} illustrates both quantitative and qualitative results for the most popular and representative \textit{Passing} and \textit{Crossing 1} scenarios.
In both, AMD and perceived Safety are strongly correlated across the eight controllers; for instance, the largest AMD of the SFM-MPPI in Passing is reflected in its highest Safety rating. The relation between CHC and Naturalness can be only noticed over single cases that stand out: in Passing, ORCA shows the highest CHC and the lowest Naturalness, while in Crossing 1 Naturalness is also affected by other factors, such as the stuttering behavior that penalized the overly cautious SFM-DWA.  
For $\mathrm{SW_s}$ is a more complete and complex metrics, hence, no consistent trend emerges compared to single aspect human scores; within each controller family, however, the SFM-based variant has both the lowest $\mathrm{SW_s}$ and the highest or equal Friendliness. 
These considerations, based on our results, show that RESONATE benchmark paves the way for more evaluation and metrics correlation studies. Most recent learning and pattern recognition methods can shed new light on human-aligned evaluation in diverse scenarios and settings of robot navigation \cite{roman2026}.

\section{Conclusions}

This paper introduces RESONATE, a reproducible real-world benchmark for social navigation based on a commercial robotic platform, a unified ROS~2 software stack, precise trajectory ground truth, and structured social interaction scenarios. The benchmark enables comparisons of eight controller configurations across eight representative scenarios and 320 physical experiments, using both quantitative and human-surveyed evaluation criteria.
The results indicate scenario-dependent rankings across methods, underscoring the importance of a categorized, structured benchmarking protocol. Nonetheless, clear results patterns have been identified to drive the research community to identify competitive baselines and best practice evaluation protocols. 
We found that human ratings of the experiments partially confirm the hypotheses of the quantitative analysis. From a general perspective, crowd-inspired models such as SFM and ORCA cannot be considered as proper standalone solutions for social navigation. Local planners alone often demonstrate more reliable, natural, and accepted trajectories. In detail, the DWA controller with Gaussian costmap achieved the highest overall reliability, whilst the version with SFM-based cost and the ORCA are considered the safest according to both proxemics and human ratings, despite their low success rate in the Narrow Turn scenario. Overall, the MPPI-based controller with SFM-based cost has been identified as the most balanced solution between robustness, navigation efficiency, and social acceptance. These findings confirm the importance of categorized real-world evaluation for assessing deployment-oriented performance beyond simulation.

Future work will extend the framework to onboard perception and learning-based methods, adopt GPU-oriented hardware, and investigate both robustness and latency issues. Nonetheless, the data collected can support new research directions to investigate correlations between quantitative and qualitative metrics for human-aligned evaluation.
Overall, RESONATE represents a practical basis for a fair, real-world comparison of socially aware navigation methods at community level, it provides accessible baselines and reproducible protocols, paving the way for future concrete development of autonomous service robots.

\backmatter

\bmhead{Acknowledgements}
This work has been developed with the contribution of the Politecnico di Torino Interdepartmental Centre for Service Robotics (PIC4SeR\footnote{\url{https://pic4ser.polito.it/}}).
This work was partially supported by the SWIch action (P.R. F.E.S.R.2021/27 - D.G.R. n.19-6962) within the EMPATHY project.

\section*{Statements and Declarations}
\bmhead{Funding}
This work was partially supported by the SWIch action (P.R. F.E.S.R. 2021/27 - D.G.R. n.~19-6962) within the EMPATHY project.

\bmhead{Competing interests}
The authors have no competing interests to disclose.

\bmhead{Ethics approval}
This study was performed in line with the principles of the Declaration of Helsinki. Approval was granted by the Ethics Committee of Politecnico di Torino (protocol n.~67359/2026).

\bmhead{Consent to participate}
Written informed consent was obtained from all individual participants included in the study.

\bmhead{Consent for publication}
The participants whose images appear in Fig.~\ref{fig:hardware} and Fig.~\ref{fig:experiment_lab} provided informed consent for the publication of these images.

\bmhead{Data availability}\label{bm:data}
The trajectories recorded in all 320 runs, with their motion-capture ground truth, and the anonymized participant ratings are publicly available on Zenodo ~\cite{resonate_dataset}.

\bmhead{Code availability}\label{bm:code}
The ROS~2 and Nav2 plugins, maps, configuration files, and evaluation pipeline used in this study are available at \url{https://github.com/PIC4SeR/RESONATE} and \url{https://github.com/PIC4SeR/social_evaluation}.

\bmhead{Supplementary information}
Online Resource 1 includes installation instructions, example trajectories, and per-scenario box plots of all metrics.

\bibliography{bibliography}

@article{gao2022evaluation,
  author  = {Yuxiang Gao and Chien-Ming Huang},
  title   = {Evaluation of Socially-Aware Robot Navigation},
  journal = {Frontiers in Robotics and AI},
  volume  = {8},
  pages   = {721317},
  year    = {2022},
  doi     = {10.3389/frobt.2021.721317}
}

@article{ros2,
    author = {Steven Macenski and Tully Foote and Brian Gerkey and Chris Lalancette and William Woodall},
    title = {Robot Operating System 2: Design, architecture, and uses in the wild},
    journal = {Science Robotics},
    volume = {7},
    number = {66},
    pages = {eabm6074},
    year = {2022},
    doi = {10.1126/scirobotics.abm6074},
    URL = {https://www.science.org/doi/abs/10.1126/scirobotics.abm6074}
}

@article{fox1997dynamic,
  author={Fox, D. and Burgard, W. and Thrun, S.},
  journal={IEEE Robotics \& Automation Magazine}, 
  title={The dynamic window approach to collision avoidance}, 
  year={1997},
  volume={4},
  number={1},
  pages={23-33},
  doi={10.1109/100.580977}
}

@article{williams2017model,
author = {Williams, Grady and Aldrich, Andrew and Theodorou, Evangelos A.},
title = {Model Predictive Path Integral Control: From Theory to Parallel Computation},
journal = {Journal of Guidance, Control, and Dynamics},
volume = {40},
number = {2},
pages = {344-357},
year = {2017},
doi = {10.2514/1.G001921},
URL = {https://doi.org/10.2514/1.G001921}
}

@article{macenski2023survey,
   title={From the desks of {ROS} maintainers: A survey of modern \& capable mobile robotics algorithms in the robot operating system 2},
   volume={168},
   ISSN={0921-8890},
   url={http://dx.doi.org/10.1016/j.robot.2023.104493},
   DOI={10.1016/j.robot.2023.104493},
   journal={Robotics and Autonomous Systems},
   publisher={Elsevier BV},
   author={Macenski, Steve and Moore, Tom and Lu, David V. and Merzlyakov, Alexey and Ferguson, Michael},
   year={2023},
   month=oct, pages={104493} 
}

@article{Francis2025guidelines,
author = {Francis, Anthony and P\'{e}rez-D’Arpino, Claudia and Li, Chengshu and Xia, Fei and Alahi, Alexandre and Alami, Rachid and Bera, Aniket and Biswas, Abhijat and Biswas, Joydeep and Chandra, Rohan and Chiang, Hao-Tien Lewis and Everett, Michael and Ha, Sehoon and Hart, Justin and How, Jonathan P. and Karnan, Haresh and Lee, Tsang-Wei Edward and Manso, Luis J. and Mirsky, Reuth and Pirk, S\"{o}ren and Singamaneni, Phani Teja and Stone, Peter and Taylor, Ada V. and Trautman, Peter and Tsoi, Nathan and V\'{a}zquez, Marynel and Xiao, Xuesu and Xu, Peng and Yokoyama, Naoki and Toshev, Alexander and Mart\'{\i}n-Mart\'{\i}n, Roberto},
title = {Principles and Guidelines for Evaluating Social Robot Navigation Algorithms},
year = {2025},
issue_date = {June 2025},
publisher = {Association for Computing Machinery},
address = {New York, NY, USA},
volume = {14},
number = {2},
url = {https://doi.org/10.1145/3700599},
doi = {10.1145/3700599},
journal = {J. Hum.-Robot Interact.},
month = feb,
articleno = {34},
numpages = {65}
}

@InProceedings{hirose2024selfi,
title={{SELFI}: Autonomous Self-Improvement with {RL} for Vision-Based Navigation around People},
author={Noriaki Hirose and Dhruv Shah and Kyle Stachowicz and Ajay Sridhar and Sergey Levine},
booktitle={8th Annual Conference on Robot Learning},
year={2024},
url={https://openreview.net/forum?id=rRpmVq6yHv}
}

@InProceedings{ORCA_paper,
author="{van den Berg}, Jur
and Guy, Stephen J.
and Lin, Ming
and Manocha, Dinesh",
editor="Pradalier, C{\'e}dric
and Siegwart, Roland
and Hirzinger, Gerhard",
title="Reciprocal n-Body Collision Avoidance",
booktitle="Robotics Research",
year="2011",
publisher="Springer Berlin Heidelberg",
address="Berlin, Heidelberg",
pages="3--19",
isbn="978-3-642-19457-3"
}

@book{hall1966hidden,
  author    = {Hall, Edward T.},
  title     = {The Hidden Dimension},
  publisher = {Doubleday \& Company, Inc.},
  address   = {Garden City, NY, USA},
  year      = {1966},
  edition   = {1st},
}

@ARTICLE{PerezRal2023,
  author={Pérez-Higueras, Noé and Otero, Roberto and Caballero, Fernando and Merino, Luis},
  journal={IEEE Robotics and Automation Letters}, 
  title={{HuNavSim}: A {ROS 2} Human Navigation Simulator for Benchmarking Human-Aware Robot Navigation}, 
  year={2023},
  month={September},
  volume={8},
  number={11},
  pages={7130-7137},
  issn={2377-3766},
  doi={10.1109/LRA.2023.3316072}}

@article{Mavrogiannis2023,
author = {Mavrogiannis, Christoforos and Baldini, Francesca and Wang, Allan and Zhao, Dapeng and Trautman, Pete and Steinfeld, Aaron and Oh, Jean},
title = {Core Challenges of Social Robot Navigation: A Survey},
year = {2023},
issue_date = {September 2023},
publisher = {Association for Computing Machinery},
address = {New York, NY, USA},
volume = {12},
number = {3},
url = {https://doi.org/10.1145/3583741},
doi = {10.1145/3583741},
journal = {J. Hum.-Robot Interact.},
month = apr,
articleno = {36},
numpages = {39}
}

@Article{huang2023hospitals,
author="Huang, Rong
and Li, Hongxiu
and Suomi, Reima
and Li, Chenglong
and Peltoniemi, Teijo",
title="Intelligent Physical Robots in Health Care: Systematic Literature Review",
journal="J Med Internet Res",
year="2023",
month="Jan",
day="18",
volume="25",
pages="e39786",
issn="1438-8871",
doi="10.2196/39786",
url="https://doi.org/10.2196/39786",
}

@INPROCEEDINGS{Blesson2023airportwheelchair,
  author={Luke, Blesson Mathews and Tripathi, Achint and Sampson, S Abraham},
  booktitle={2023 International Conference on Next Generation Electronics (NEleX)}, 
  title={Autonomous Wheelchair Robot for Enhanced Airport Accessibility}, 
  year={2023},
  volume={},
  number={},
  pages={1-6},
  doi={10.1109/NEleX59773.2023.10420850}}

@article{Helbing1995SocialFM,
  author  = {Dirk Helbing and P{\'e}ter Moln{\'a}r},
  title   = {Social force model for pedestrian dynamics},
  journal = {Physical Review E},
  volume  = {51},
  number  = {5},
  pages   = {4282--4286},
  year    = {1995},
  doi     = {10.1103/PhysRevE.51.4282}
}

@article{biswas2022socnavbench,
author = {Biswas, Abhijat and Wang, Allan and Silvera, Gustavo and Steinfeld, Aaron and Admoni, Henny},
title = {SocNavBench: A Grounded Simulation Testing Framework for Evaluating Social Navigation},
year = {2022},
issue_date = {September 2022},
publisher = {Association for Computing Machinery},
address = {New York, NY, USA},
volume = {11},
number = {3},
url = {https://doi.org/10.1145/3476413},
doi = {10.1145/3476413},
journal = {J. Hum.-Robot Interact.},
month = jul,
articleno = {26},
numpages = {24}
}

@ARTICLE{karnan2022scand,
  author={Karnan, Haresh and Nair, Anirudh and Xiao, Xuesu and Warnell, Garrett and Pirk, Sören and Toshev, Alexander and Hart, Justin and Biswas, Joydeep and Stone, Peter},
  journal={IEEE Robotics and Automation Letters}, 
  title={Socially CompliAnt Navigation Dataset ({SCAND}): A Large-Scale Dataset of Demonstrations for Social Navigation}, 
  year={2022},
  volume={7},
  number={4},
  pages={11807-11814},
  doi={10.1109/LRA.2022.3184025}}

@article{kobayashi2022robot,
	title = {Robot {Navigation} {Based} on {Predicting} of {Human} {Interaction} and its {Reproducible} {Evaluation} in a {Densely} {Crowded} {Environment}},
	volume = {14},
	issn = {1875-4805},
	url = {https://doi.org/10.1007/s12369-021-00791-9},
	doi = {10.1007/s12369-021-00791-9},
	number = {2},
	journal = {International Journal of Social Robotics},
	author = {Kobayashi, Yuichi and Sugimoto, Takeshi and Tanaka, Kazuhito and Shimomura, Yuki and Arjonilla Garcia, Francisco J. and Kim, Chyon Hae and Yabushita, Hidenori and Toda, Takahiro},
	month = mar,
	year = {2022},
	pages = {373--387},
}

@inproceedings{sarl_crowd_paper,
  author={Chen, Changan and Liu, Yuejiang and Kreiss, Sven and Alahi, Alexandre},
  booktitle={2019 International Conference on Robotics and Automation (ICRA)}, 
  title={Crowd-Robot Interaction: Crowd-Aware Robot Navigation With Attention-Based Deep Reinforcement Learning}, 
  year={2019},
  volume={},
  number={},
  pages={6015-6022},
  doi={10.1109/ICRA.2019.8794134}}

@inproceedings{CADRL_paper,
  author={Chen, Yu Fan and Liu, Miao and Everett, Michael and How, Jonathan P.},
  booktitle={2017 IEEE International Conference on Robotics and Automation (ICRA)}, 
  title={Decentralized non-communicating multiagent collision avoidance with deep reinforcement learning}, 
  year={2017},
  volume={},
  number={},
  pages={285-292},
  doi={10.1109/ICRA.2017.7989037}}

@article{SICNav-Diffusion,
   title={{SICNav-Diffusion}: Safe and Interactive Crowd Navigation With Diffusion Trajectory Predictions},
   volume={10},
   ISSN={2377-3774},
   url={http://dx.doi.org/10.1109/LRA.2025.3585713},
   DOI={10.1109/lra.2025.3585713},
   number={9},
   journal={IEEE Robotics and Automation Letters},
   publisher={Institute of Electrical and Electronics Engineers (IEEE)},
   author={Samavi, Sepehr and Lem, Anthony and Sato, Fumiaki and Chen, Sirui and Gu, Qiao and Yano, Keijiro and Schoellig, Angela P. and Shkurti, Florian},
   year={2025},
   month=sep, pages={8738–8745} }

@article{rudenko2020thor,
  author={Rudenko, Andrey and Kucner, Tomasz P. and Swaminathan, Chittaranjan S. and Chadalavada, Ravi T. and Arras, Kai O. and Lilienthal, Achim J.},
  journal={IEEE Robotics and Automation Letters}, 
  title={{TH\"OR}: Human-Robot Navigation Data Collection and Accurate Motion Trajectories Dataset}, 
  year={2020},
  volume={5},
  number={2},
  pages={676-682},
  doi={10.1109/LRA.2020.2965416}}

@article{schreiter2025thormagni,
title     = {{TH\"OR-MAGNI}: A large-scale indoor motion capture recording of human movement and robot interaction},
author = {Tim Schreiter and Tiago Rodrigues de Almeida and Yufei Zhu and Eduardo Gutierrez Maestro and Lucas Morillo-Mendez and Andrey Rudenko and Luigi Palmieri and Tomasz P Kucner and Martin Magnusson and Achim J Lilienthal},

journal = {The International Journal of Robotics Research},
volume = {44},
number = {4},
pages = {568-591},
year = {2025},
doi = {10.1177/02783649241274794},
URL = { 
        https://doi.org/10.1177/02783649241274794
},
}

@article{tsoi2022sean2,
  title     = {{SEAN} 2.0: Formalizing and Generating Social Situations for Robot Navigation},
  author={Tsoi, Nathan and Xiang, Alec and Yu, Peter and Sohn, Samuel S. and Schwartz, Greg and Ramesh, Subashri and Hussein, Mohamed and Gupta, Anjali W. and Kapadia, Mubbasir and Vázquez, Marynel},
  journal={IEEE Robotics and Automation Letters}, 
  year={2022},
  volume={7},
  number={4},
  pages={11047-11054},
  doi={10.1109/LRA.2022.3196783}}

@article{kastner2022arenabench,
  author={K{\"a}stner, Linh and Bhuiyan, Teham and L{\^e}, Tuan Anh and Treis, Elias and Cox, Johannes and Meinardus, Boris and Kmiecik, Jacek and Carstens, Reyk and Pichel, Duc and Fatloun, Bassel and Khorsandi, Niloufar and Lambrecht, Jens},
  journal={IEEE Robotics and Automation Letters}, 
  title     = {{Arena-Bench}: A Benchmarking Suite for Obstacle Avoidance Approaches in Highly Dynamic Environments},
  year={2022},
  volume={7},
  number={4},
  pages={9477-9484},
  doi={10.1109/LRA.2022.3190086}}

@misc{resonate_dataset,
  author    = {Trepella, Stefano and Ostuni, Andrea and Martini, Mauro and Borgini, Irene and Chiaberge, Marcello},
  title     = {{RESONATE}: real-world social navigation benchmark dataset},
  publisher = {Zenodo},
  year      = {2026},
  doi       = {10.5281/zenodo.23763544}
}

@INPROCEEDINGS{roman2026,
  author={Trepella, Stefano and Martini, Mauro and Ostuni, Andrea and Pérez-Higueras, Noé and Caballero, Fernando and Merino, Luis and Chiaberge, Marcello},
  booktitle={2026 IEEE 35th International Conference on Robot and Human Interactive Communication (RO-MAN)}, 
  title={Metrics vs Surveys: An Analysis for Human-Aligned Benchmarking in Social Robot Navigation}, 
  year={2026},
  volume={},
  number={},
  pages={2483-2489},
  doi={10.1109/RO-MAN72286.2026.11714815}}


\end{document}